\documentclass[11pt]{article}
\pdfoutput=1
\usepackage{ACL2023}

\usepackage{times}
\usepackage{latexsym}

\usepackage[T1]{fontenc}

\usepackage[utf8]{inputenc}

\usepackage{microtype}

\usepackage{inconsolata}

\usepackage{multirow}
\usepackage{graphicx}
\usepackage{subfigure}
\usepackage{booktabs}
\usepackage[capitalise,noabbrev]{cleveref}
\usepackage{adjustbox}
\usepackage{comment}
\usepackage{makecell}
\usepackage{xcolor}

\newcommand{\mask}{\texttt{[Mask]}}

\title{Do Multilingual Language Models Think Better in English?}

\author{Julen Etxaniz$^{1}$ \quad Gorka Azkune$^{1}$ \quad Aitor Soroa$^{1}$ \quad Oier Lopez de Lacalle$^{1}$ \quad Mikel Artetxe$^{2}$ \\
$^{1}$HiTZ Center, University of the Basque Country UPV/EHU \qquad $^{2}$Reka AI \\
\texttt{\{julen.etxaniz,gorka.azcune,a.soroa,oier.lopezdelacalle\}@ehu.eus} \quad \texttt{mikel@reka.ai} \\
}

\begin{document}
\maketitle
\begin{abstract}
Translate-test is a popular technique to improve the performance of multilingual language models. This approach works by translating the input into English using an external machine translation system, and running inference over the translated input. However, these improvements can be attributed to the use of a separate translation system, which is typically trained on large amounts of parallel data not seen by the language model. In this work, we introduce a new approach called self-translate, which overcomes the need of an external translation system by leveraging the few-shot translation capabilities of multilingual language models. Experiments over 5 tasks show that self-translate consistently outperforms direct inference, demonstrating that language models are unable to leverage their full multilingual potential when prompted in non-English languages. Our code is available at \url{https://github.com/juletx/self-translate}.
\end{abstract}

\section{Introduction}
\label{sec:introduction}

Multilingual autoregressive language models like XGLM \citep{lin-etal-2022-shot}, BLOOM \citep{workshop2023bloom} and PaLM \citep{chowdhery2022palm,anil2023palm2} have shown impressive capabilities on many tasks and languages. However, performance is usually lower for non-English languages, especially for low-resource ones \citep{ahuja2023mega}. A common approach to mitigate this problem is to use translate-test, where the test data is translated into English using an external Machine Translation (MT) system, and then fed into the model. While primarily explored in the traditional pretrain/finetune paradigm \citep{ponti2021modelling,artetxe2023revisiting}, early evidence has shown that translate-test can also bring sizeable improvements for few-shot learning with autoregressive language models \citep{shi2022language}.

\begin{figure*}[t]
\centering
\includegraphics[width=0.91\textwidth]{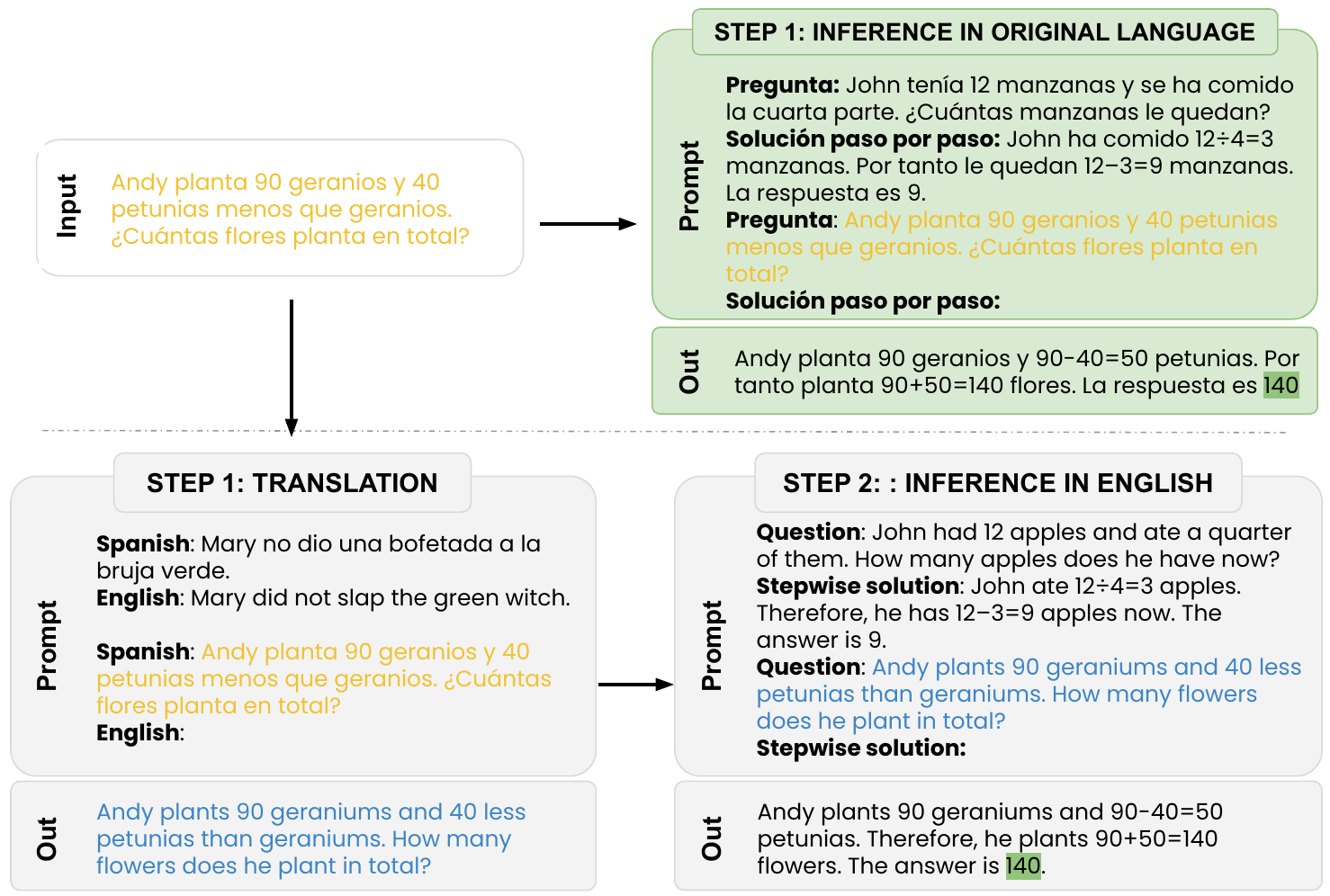}
\caption{\textbf{Direct inference (top) vs. self-translate (bottom).} In direct inference (standard) the task is solved by prompting the model in the original language. In self-translate (proposed), we first translate the input into English by prompting the same model, and then solve the task in English.}
\label{fig:method}
\end{figure*}

However, translate-test relies on a separate MT system, which is usually trained on large amounts of parallel data not seen by the primary model.
In this paper, we investigate if the improvements from translate-test are solely due to the use of additional resources.
To answer this question, we propose a new approach called self-translate, which leverages the few-shot translation capabilities of autoregressive language models \citep{vilar-etal-2023-prompting} instead of using an external system. More concretely, we prompt multilingual models to translate the input into English, and then feed the translated input to the same model to solve the task (Figure \ref{fig:method}). 

\begin{figure}[t]
\centering
\includegraphics[width=\linewidth]{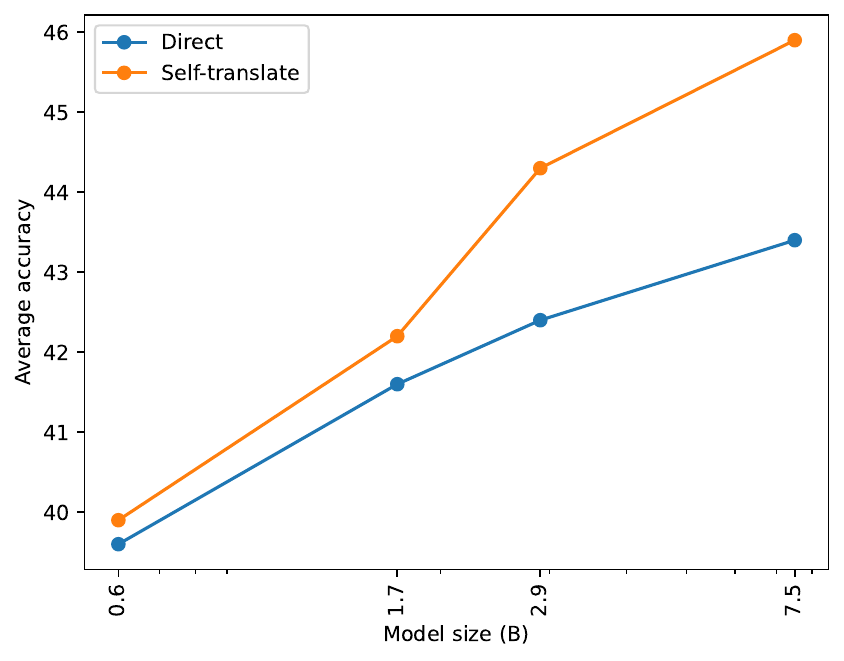}
\caption{\textbf{XGLM results (average accuracy).} We show that self-translate (using the model itself to translate the input into English) works better than using the original input in the non-English language.}
\label{fig:xglm_avg}
\end{figure}

As shown in Figure \ref{fig:xglm_avg}, we find that self-translate works better than solving the task directly in the original language. This demonstrates that multilingual language models are unable to leverage their full potential when prompted in non-English languages. We find this phenomenon to be consistent across tasks, and more prominent for large models and high-resource languages. All in all, our work reveals an important limitation of multilingual language models, and prompts for future work to unleash their full potential without the need of intermediate inference steps.

\section{Experimental settings}
\label{sec:experiments}

We next describe our experimental design, and report additional details in \cref{app:experimental_details}.

\begin{table*}[t!]
\centering
\small
\begin{adjustbox}{max width=\linewidth}
\begin{tabular}{crllllll|l}
\toprule
Model & Size & Method & \multicolumn{1}{c}{XStoryC} & \multicolumn{1}{c}{XCOPA} & \multicolumn{1}{c}{XNLI} & \multicolumn{1}{c}{PAWS-X} & \multicolumn{1}{c}{MGSM} & \multicolumn{1}{|c}{Avg} \\ \midrule
\multirow{9.5}{*}{XGLM}
& \multirow{2}{*}{0.6B}
& Direct &          \textbf{53.5} &   \textbf{54.9} &  39.4 &   48.4 & \textbf{1.7} & 39.6 \\
&& Self-translate &          52.8 \textsubscript{(-0.8)} &   53.4 \textsubscript{(-1.5)} &  \textbf{41.5 \textsubscript{(+2.1)}} &   \textbf{50.6 \textsubscript{(+2.2)}} & 1.4 \textsubscript{(-0.3)} & \textbf{39.9 \textsubscript{(+0.3)}} \\
\cmidrule{2-9}
& \multirow{2}{*}{1.7B}
& Direct &          \textbf{56.5} &   57.1 &  41.9 &   \textbf{50.7} & \textbf{1.7} & 41.6 \\
&& Self-translate &          55.9 \textsubscript{(-0.6)} &  \textbf{58.4  \textsubscript{(+1.3)}} &  \textbf{44.9 \textsubscript{(+3.0)}} &   50.2 \textsubscript{(-0.5)} & \textbf{1.7 \textsubscript{(+0.0)}} & \textbf{42.2 \textsubscript{(+0.6)}} \\
\cmidrule{2-9}
& \multirow{2}{*}{2.9B}
& Direct &          \textbf{58.2} &   58.5 &  43.0 &   50.8 & 1.4 & 42.4 \\
&& Self-translate &          \textbf{58.2 \textsubscript{(+0.0)}} &   \textbf{62.5 \textsubscript{(+4.0)}} &  \textbf{46.2 \textsubscript{(+3.2)}} &   \textbf{53.2 \textsubscript{(+2.4)}} & \textbf{1.6 \textsubscript{(+0.2)}} & \textbf{44.3 \textsubscript{(+1.9)}} \\
\cmidrule{2-9}
& \multirow{2}{*}{7.5B}
& Direct &          59.9 &   60.6 &  44.0 &   51.6 & \textbf{0.8} & 43.4 \\
&& Self-translate &          \textbf{60.9 \textsubscript{(+1.0)}} &   \textbf{64.4 \textsubscript{(+3.8)}} &  \textbf{48.9 \textsubscript{(+4.9)}} &   \textbf{55.4 \textsubscript{(+3.8)}} & 0.1  \textsubscript{(-0.7)} & \textbf{45.7 \textsubscript{(+2.3)}} \\

\midrule

\multirow{7}{*}{LLaMA}
& \multirow{2}{*}{7B}
& Direct &          53.6 &   53.9 &  37.1 &   53.2 &   5.0 & 40.6 \\
&&  Self-translate &          \textbf{55.8 \textsubscript{(+2.2)}} &   \textbf{54.9 \textsubscript{(+1.0)}} &  \textbf{43.0 \textsubscript{(+5.9)}} &   \textbf{57.0 \textsubscript{(+3.8)}} &   \textbf{6.1 \textsubscript{(+1.1)}} & \textbf{43.4 \textsubscript{(+2.8)}} \\
\cmidrule{2-9}
& \multirow{2}{*}{13B}
& Direct &          54.8 &   54.7 &  34.2 &   49.5 &   7.4 & 40.1 \\
&&  Self-translate &          \textbf{57.7 \textsubscript{(+2.9)}} &   \textbf{56.5 \textsubscript{(+1.8)}} &  \textbf{35.1 \textsubscript{(+0.9)}} &   \textbf{52.1 \textsubscript{(+2.6)}} &  \textbf{10.0 \textsubscript{(+2.6)}} & \textbf{42.3 \textsubscript{(+2.2)}} \\
\cmidrule{2-9}
& \multirow{2}{*}{30B}
& Direct &          56.7 &   55.2 &  37.0 &   50.9 &  15.5 & 43.1 \\
&&  Self-translate &          \textbf{59.0 \textsubscript{(+2.3)}} &   \textbf{58.4 \textsubscript{(+3.2)}} &  \textbf{43.5 \textsubscript{(+6.5)}} &   \textbf{55.6 \textsubscript{(+4.7)}} &  \textbf{16.3 \textsubscript{(+0.8)}} & \textbf{46.6 \textsubscript{(+3.5)}} \\
\bottomrule
\end{tabular}
\end{adjustbox}
\caption{\textbf{Main results (accuracy).} Task performance in terms of accuracy for different sizes of XGLM and LLaMA, using \textbf{direct} inference and \textbf{self-translate}. The last column shows the average accuracy over all tasks.
We highlight the best results for each model and task in bold.%
}
\label{tab:main_results}
\end{table*}

\paragraph{Models.} We experiment with 7 models from 2 families: the 564M, 1.7B, 2.9B and 7.5B models from \textbf{XGLM} \citep{lin-etal-2022-shot}, and the 7B, 13B and 30B models from \textbf{LLaMA} \cite{touvron2023llama}. XGLM has a multilingual focus and covers many languages, but is smaller in size and lags behind recent models in English. In contrast, LLaMA is primarily trained on English and is much stronger in this language, while also showing some multilingual capabilities. \cref{app:additional_results} reports additional results for BLOOM \citep{workshop2023bloom}, LLaMA 2 \citep{touvron2023llama2}, OpenLLaMA \citep{openlm2023openllama}, OpenLLaMA V2 \citep{openlm2023openllama}, Redpajama \citep{together2023redpajama} and PolyLM \citep{wei2023polylm}.

\paragraph{Methods.} As shown in Figure \ref{fig:method}, we compare two methods for each model: \textbf{direct} inference, where we feed the original (non-English) input to the model, and \textbf{self-translate}, where we first translate the input into English using the model itself, and then feed this translated input to the same model to solve the task. For translation, we do 4-shot prompting using examples from the FLORES-200 dataset \cite{costa2022no}, prepending each sentence with its corresponding language name. We select the first sentences from the development set, skipping those that are longer than 100 characters. %
We use greedy decoding, and translate each field in the input (e.g., the premise and hypothesis in XNLI) separately. For analysis, we additionally compare self-translate to using an external state-of-the-art \textbf{MT} system. To that end, we use the 3.3B NLLB-200 model \cite{costa2022no}.

\paragraph{Evaluation.} We use the following tasks for evaluation: \textbf{XCOPA} \cite{ponti2020xcopa}, a common sense reasoning task in 11 languages; \textbf{XStoryCloze} \cite{lin-etal-2022-shot}, a common sense reasoning task in 11 languages; \textbf{XNLI} \cite{conneau2018xnli}, a natural language inference task in 15 languages; \textbf{PAWS-X} \cite{yang2019paws}, a paraphrase identification task in 7 languages; and \textbf{MGSM} \cite{shi2022language}, a mathematical reasoning task with grade school problems in 11 languages. For MGSM, we do 8-shot evaluation with a chain-of-thought prompt, and extract the answer using a regular expression. The rest of the tasks are not generative, so we feed each candidate in a zero-shot fashion and pick the one with the highest probability.

\section{Results}
\label{sec:results}

\begin{figure*}[ht]
\centering
\subfigure{\includegraphics[width=0.48\textwidth]{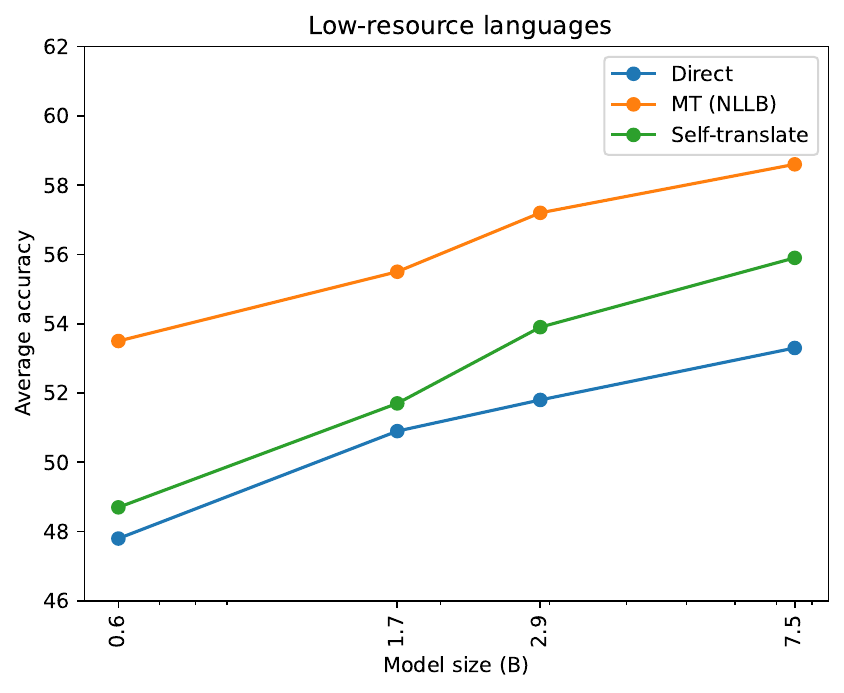}}
\vspace{-0.5cm}
\subfigure{\includegraphics[width=0.48\textwidth]{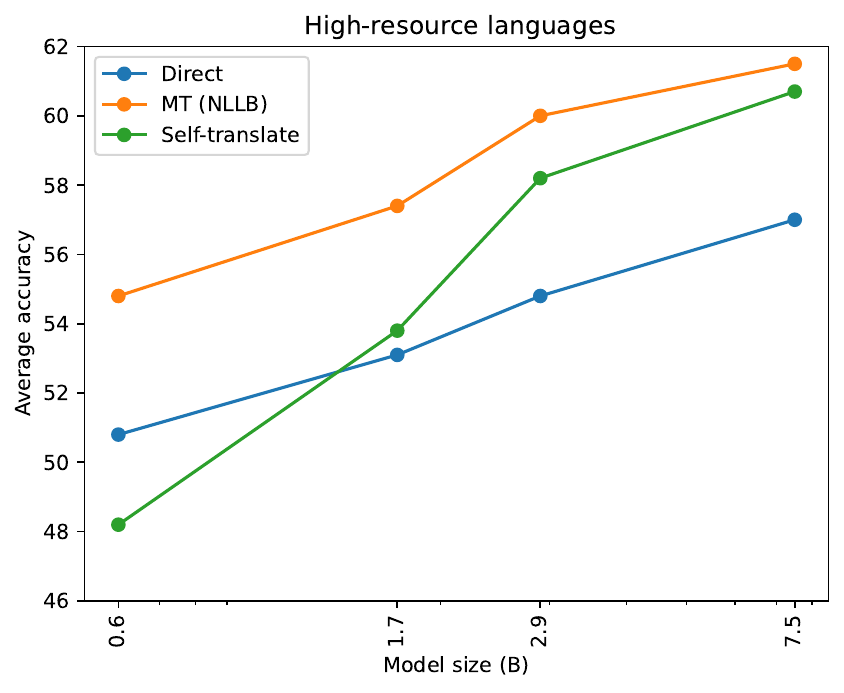}}
\subfigure{\includegraphics[width=0.48\textwidth]{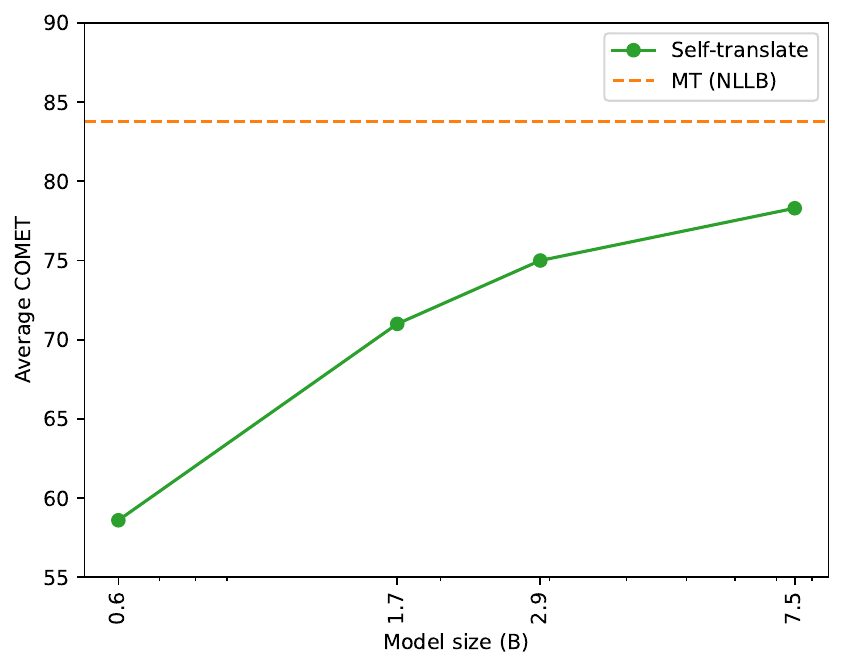}}
\vspace{-0.4cm}
\subfigure{\includegraphics[width=0.48\textwidth]{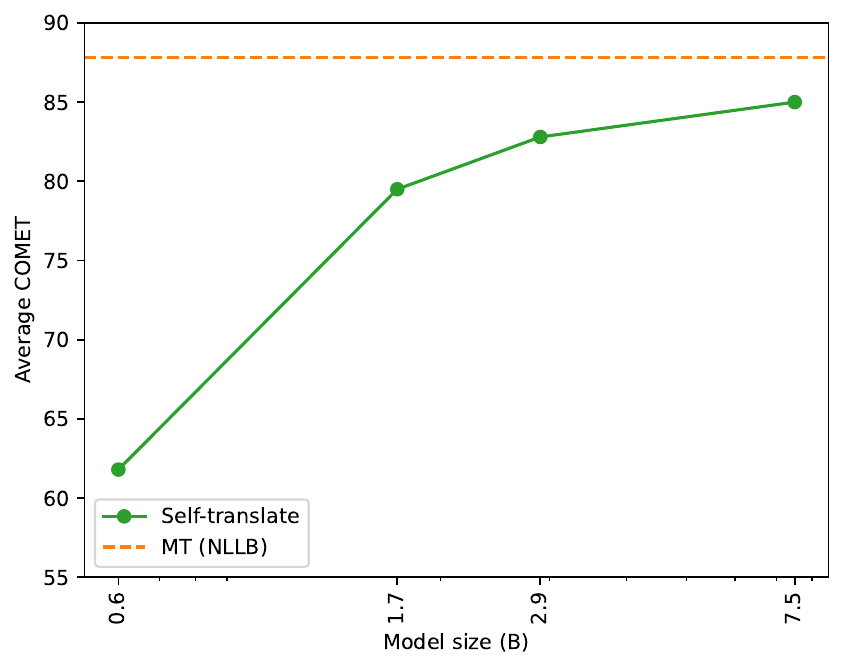}}
\caption{\textbf{Downstream (top) and MT (bottom) performance, grouped by low-resource (left) and high-resources (right) languages.} For downstream, we report average accuracy over XStoryCloze, XCOPA and XNLI, which have the most language variety. Low- and high-resource languages follow \citet{lin-etal-2022-shot}, merging the low and ex-low categories. For MT, we report COMET \cite{rei2022comet}, using the target language text for each field in those datasets as the source, and the English text as the reference.%
}
\label{fig:xglm_resources}
\end{figure*}

Table \ref{tab:main_results} reports our main results, and Figure \ref{fig:xglm_avg} visualizes the average accuracy of XGLM as a function of scale. \cref{fig:xglm_resources} compares the downstream performance and translation quality of self-translate and NLLB, grouped by low-resource and high-resource languages.
Additional results are reported in Appendix \ref{app:additional_results}. We next summarize our main findings:

\paragraph{Self-translate outperforms direct inference.} %
We find that self-translate works better than direct inference in average for all models. The results are also consistent across tasks, with only a few exceptions for the smaller XGLM models. This proves that multilingual language models are more capable than immediately obvious in non-English languages, but unveiling their full potential requires performing intermediate steps.

\paragraph{Multilingual language models do transfer capabilities across languages.} One possible explanation for the previous finding is that language models acquire capabilities separately for each language, without any effective cross-lingual transfer.
However, a closer comparison of LLaMA and XGLM refutes this hypothesis. %
In particular, we observe that LLaMA is much better than XGLM in MGSM despite being worse in other tasks. This is because MGSM is an emergent task \citep{wei2022emergent}, and XGLM, being smaller and less capable, obtains near 0 accuracy. In contrast, LLaMA is more capable at solving math word problems, and it is able to leverage this capability even if prompted in other languages. The superior performance of self-translate shows that this cross-lingual transfer is not fully effective, but our results suggest that it does happen to a large extent.

\paragraph{Self-translate is more effective for high-resource languages and large models.}
Figure \ref{fig:xglm_avg} shows that the gap between self-translate and direct inference gets larger at scale. Similarly, as shown by Table \ref{tab:main_results}, it is the largest LLaMA model that obtains the biggest absolute gains over direct inference. At the same time, Figure \ref{fig:xglm_resources} (top) shows that the effect of scale is bigger for high-resource languages and, for the largest model sizes, high-resource languages benefit more from self-translate than low-resource languages. This suggests that the effectiveness of self-translate is not explained by the limited capacity of smaller models, and can be expected to increase at scale.

\paragraph{MT outperforms self-translate, but the gap narrows at scale.} As shown by Figure \ref{fig:xglm_resources} (top), NLLB performs better than self-translate, meaning that it can still be beneficial to use an external MT system. However, the gap narrows at scale, as the translation capabilities of the largest models approach NLLB (Figure \ref{fig:xglm_resources}, bottom). Given the recent claims that state-of-the-art multilingual language models are competitive with traditional MT systems \cite{vilar-etal-2023-prompting, hendy2023good}, this suggests that stronger language models would not require an external MT system for best results.

\section{Related work}

Translate-test is a strong baseline in the traditional pretrain/finetune paradigm \citep{ponti2021modelling,artetxe2023revisiting}. Early evidence shows that it is also effective for prompting autoregressive language models \citep{lin-etal-2022-shot,shi2022language}, as these models have irregular performance depending on the input language \cite{bang2023multitask}. Recent work has shown that multilingual language models are good translators \cite{zhang2023prompting,hendy2023good,vilar-etal-2023-prompting}, which our approach exploits to replace the external MT system in translate-test. Concurrent to our work, \citet{huang2023not} propose a more complex prompting method that involves translating the input, but they only experiment with proprietary models and do not study the role of translation in isolation. Finally, \citet{reid-artetxe-2023-role} show that using synthetic parallel data from unsupervised MT can improve the performance of multilingual models, but they focus on pretraining seq2seq models.

\section{Conclusion}
\label{sec:conclusion}

We have proposed a new method called self-translate, where we use a multilingual language model to translate the test data into English, and then feed the translated data to the same model to solve the task. Self-translate consistently outperfoms the standard direct inference approach, which directly feeds the test data in the original language. Our approach does not involve any additional data or training, showing that language models are not able to leverage their full multilingual potential when prompted in non-English languages. In the future, we would like to explore training methods to mitigate this issue without the need of intermediate inference steps.

\section*{Limitations}
\label{sec:limitations}

Despite consistently outperforming direct inference, self-translate is substantially slower due to the cost of the translation step.

Our goal was to study a fundamental limitation of multilingual language models, and we decided to use base models to that end. In practice, instruction-tuned models would remove the need for few-shot prompts and make self-translate more efficient, as well as enabling to translate and solve the task in a single step. %

Finally, all the datasets that we use were created through (human) translation, which can result in evaluation artifacts for methods involving machine translation \citep{artetxe-etal-2020-translation}. A more realistic scenario would be to use datasets that are natively written in different languages, but such datasets are scarce and not standard for evaluating autoregressive language models.

\section*{Acknowledgements}
Julen is funded by a PhD grant from the Basque Government (PRE\_2022\_1\_0047). This work is partially supported by projects founded by MCIN/AEI/10.13039/501100011033 and European Union NextGeneration EU/PRTR (DeepR3 TED2021-130295B-C31, AWARE TED2021-131617B-I00, and DeepKnowledge PID2021-127777OB-C21), and the Basque Government (IXA excellence research group IT1570-22, IKER-GAITU 11:4711:23:410:23/0808 and NEL-GAITU).

\bibliography{custom}
\bibliographystyle{acl_natbib}

\appendix

\section{Experimental details} \label{app:experimental_details}

In this section, we report additional experimental details that cover the evaluation library, task descriptions and prompts.

\subsection{Evaluation library}

We use LM Evaluation Harness \cite{eval-harness} for evaluation. There were no multilingual tasks in the library, so we decided to add them so that our results can be replicated and extended to more models. For self-translate and MT, we define another evaluation task that uses a different dataset format. We created a fork of the evaluation library that includes these additional tasks at \url{https://github.com/juletx/lm-evaluation-harness/tree/translation}. All the translations generated with self-translate and MT are available at \url{https://huggingface.co/juletxara}.

\subsection{Prompts}

For self-translate and MT, we used the same English prompts used in XGLM to evaluate most tasks (Table \ref{tab:english_prompts}). For direct inference, we use multilingual prompts, which are already available in some datasets (e.g. MGSM). When multilingual prompts are not available, we create them by translating English prompts to each language, using Google Translate. Note that this is suboptimal because translations are generally not as good as native prompts. Another option would be to always use English prompts, but this is also unnatural because it adds English tokens in the middle of other languages. All the multilingual prompts are available in the evaluation library above.

\begin{table*}[ht]
    \centering
    \begin{adjustbox}{max width=\linewidth}
    \begin{tabular}{c|l|c}
        \toprule
        Task & Template & Candidate Verbalizer  \\
        \midrule
        \multirow{2}{*}{XCOPA} & \textit{cause}: \texttt{\{Sentence 1\}} because \mask & \multirow{2}{*}{Identity} \\
         & \textit{effect}: \texttt{\{Sentence 1\}} therefore \mask & \\
         \midrule
        XStoryCloze & \texttt{\{Context\}} \mask & Identity \\
        \midrule
        XNLI & \texttt{\{Sentence 1\}}, right? \mask, \texttt{\{Sentence 2\}} & \makecell{\textit{Entailment}: Yes | \textit{Neutral}: Also | \textit{Contradiction}: No}\\
        \midrule
        PAWS-X & \texttt{\{Sentence 1\}}, right? \mask, \texttt{\{Sentence 2\}} & \makecell{\textit{True}: Yes | \textit{False}: No}\\
        \midrule
        MGSM & Question: \texttt{\{Question\}} Step-by-Step Answer: & None \\
        \bottomrule
    \end{tabular}
    \end{adjustbox}
    \caption{\textbf{Handcrafted English prompts for multilingual tasks.} The identity function maps each candidate choice to itself. In the case of MGSM there is no verbalizer, because the model generates an answer that is extracted with a regular expression.}
    \label{tab:english_prompts}
\end{table*}

\section{Additional results} \label{app:additional_results}

In this section, we report additional results that cover direct vs. self-translate, self-translate vs. MT, results by language and translation metrics.

\subsection{Direct vs. self-translate}
\label{sec:more_models}

We include additional direct vs. self-translate results for \textbf{BLOOM} \citep{workshop2023bloom}, \textbf{LLaMA 2} \citep{touvron2023llama2}, \textbf{OpenLLaMA} \citep{openlm2023openllama}, \textbf{OpenLLaMA V2} \citep{openlm2023openllama}, \textbf{Redpajama} \citep{together2023redpajama} and \textbf{PolyLM} \citep{wei2023polylm}. Similar to XGLM, BLOOM has a multilingual focus and covers many languages. The rest of the models are similar to LLaMA, which is primarily trained on English and is much stronger in this language, while also showing some multilingual capabilities. \cref{tab:main_results_direct_self} shows the results as accuracy of the \textbf{direct} and \textbf{self-translate} methods in all tasks for different models and sizes. Results resemble the ones obtained by XGLM and LLaMA in the main results, so we can conclude that self-translate is consistent across different models.

\begin{table*}[t]
\centering
\small
\begin{adjustbox}{max width=\linewidth}
\begin{tabular}{crllllll|l}
\toprule
Model & Size & Method & \multicolumn{1}{c}{XStoryC} & \multicolumn{1}{c}{XCOPA} & \multicolumn{1}{c}{XNLI} & \multicolumn{1}{c}{PAWS-X} & \multicolumn{1}{c}{MGSM} & \multicolumn{1}{|c}{Avg} \\ \midrule
\multirow{9.5}{*}{BLOOM}
& \multirow{2}{*}{0.6B} &         Direct &          \textbf{52.9} &   \textbf{54.0} &  36.6 &   \textbf{49.3} &   \textbf{1.7} & 38.9 \\
& & Self-translate &          \textbf{52.9} &   51.0 &  \textbf{41.4} &   48.4 &   1.5 & \textbf{39.0} \\ \cmidrule{2-9}
 &   \multirow{2}{*}{1.7B} &         Direct &          55.2 &   \textbf{55.1} &  39.2 &   47.0 &   \textbf{2.3} & 39.8 \\
 &    & Self-translate &          \textbf{55.5} &   54.7 &  \textbf{41.9} &   \textbf{48.0} &   1.8 & \textbf{40.4} \\ \cmidrule{2-9}
 &   \multirow{2}{*}{3.0B} &         Direct &          56.4 &   56.1 &  39.8 &   49.4 &   2.0 & 40.7 \\
 &    & Self-translate &          \textbf{57.2} &   \textbf{56.7} &  \textbf{44.1} &   \textbf{52.1} &   \textbf{2.1} & \textbf{42.4} \\ \cmidrule{2-9}
 &   \multirow{2}{*}{7.1B} &         Direct &          58.2 &   56.9 &  40.7 &   50.2 &   \textbf{3.2} & 41.8 \\
 &   & Self-translate &          \textbf{59.3} &   \textbf{59.7} &  \textbf{45.4} &   \textbf{54.4} &   3.1 & \textbf{44.4} \\  \midrule
 \multirow{4.5}{*}{LLaMA 2}
& \multirow{2}{*}{7B} &         Direct &          55.6 &   56.7 &  39.2 &   57.9 &   1.8 & 42.2 \\
 &   & Self-translate &          \textbf{57.8} &   \textbf{59.3} &  \textbf{47.6} &   \textbf{61.3} &   \textbf{7.2} & \textbf{46.6} \\  \cmidrule{2-9}
& \multirow{2}{*}{13B} &         Direct &          57.2 &   58.2 &  39.8 &   52.4 &  13.2 & 44.2 \\
 &   & Self-translate &          \textbf{59.9} &   \textbf{61.3} &  \textbf{46.0} &   \textbf{55.2} &  \textbf{19.2} & \textbf{48.3} \\  \midrule
  \multirow{4.5}{*}{RedPajama}
& \multirow{2}{*}{3B} &         Direct &          51.4 &   53.0 &  36.3 &   52.6 &   1.1 & 38.9 \\
 &    & Self-translate &          \textbf{52.3} &   \textbf{53.1} &  \textbf{41.8} &   \textbf{56.8} &   \textbf{1.4} & \textbf{41.1} \\  \cmidrule{2-9}
 &   \multirow{2}{*}{7B} &         Direct &          53.3 &   52.5 &  38.2 &   54.5 &   2.0 & 40.1 \\
 &    & Self-translate &          \textbf{53.9} &   \textbf{55.2} &  \textbf{42.6} &   \textbf{57.4} &   \textbf{3.2} & \textbf{42.5} \\  \midrule
 \multirow{6.5}{*}{OpenLLaMA}
& \multirow{2}{*}{3B} &         Direct &          51.0 &   52.4 &  35.7 &   48.4 &   1.1 & 37.7 \\
 &    & Self-translate &          \textbf{53.4} &   \textbf{52.5} &  \textbf{39.7} &   \textbf{53.1} &   \textbf{1.9} & \textbf{40.1} \\ \cmidrule{2-9}
& \multirow{2}{*}{7B} &         Direct &          52.4 &   52.9 &  37.0 &   51.8 &   1.9 & 39.2 \\
 &   & Self-translate &          \textbf{55.5} &   \textbf{53.9} &  \textbf{43.1} &   \textbf{56.9} &   \textbf{3.6} & \textbf{42.6} \\ \cmidrule{2-9}
& \multirow{2}{*}{13B} &         Direct &          53.8 &   54.0 &  38.6 &   52.7 &   3.5 & 40.5 \\
 &   & Self-translate &          \textbf{55.4} &   \textbf{56.0} &  \textbf{44.2} &   \textbf{58.0} &   \textbf{5.3} & \textbf{43.8} \\ \midrule
 \multirow{4.5}{*}{OpenLLaMA V2}
& \multirow{2}{*}{3B} &         Direct &          52.2 &   53.7 &  36.8 &   49.0 &   2.2 & 38.8 \\
 &    & Self-translate &          \textbf{54.5} &   \textbf{55.6} &  \textbf{43.4} &   \textbf{52.8} &   \textbf{3.0} & \textbf{41.9} \\ \cmidrule{2-9}
& \multirow{2}{*}{7B} &         Direct &          53.9 &   54.4 &  38.2 &   52.3 &   3.6 & 40.5 \\
 &    & Self-translate &          \textbf{55.7} &   \textbf{56.9} &  \textbf{44.6} &   \textbf{56.2} &   \textbf{5.7} & \textbf{43.8} \\ \midrule
 \multirow{4.5}{*}{PolyLM}
& \multirow{2}{*}{1.7B} &         Direct &          51.8 &   \textbf{54.3} &  37.4 &   48.2 &   1.4 & 38.6 \\
 &    & Self-translate &          \textbf{52.6} &   53.2 &  \textbf{40.6} &   \textbf{49.4} &   \textbf{1.6} & \textbf{39.5} \\ \cmidrule{2-9}
& \multirow{2}{*}{13B} &         Direct &          56.3 &   58.9 &  41.4 &   55.0 &   4.4 & 43.2 \\
 &   & Self-translate &          \textbf{57.4} &   \textbf{60.4} &  \textbf{45.6} &   \textbf{57.3} &   \textbf{5.3} & \textbf{45.2} \\
\bottomrule
\end{tabular}
\end{adjustbox}
\caption{\textbf{Direct vs. self-translate.} Task accuracy for different sizes of BLOOM, OpenLLaMA, OpenLLaMA V2, Redpajama and PolyLM, using direct inference and self-translate. The last column shows the average accuracy over all tasks. We highlight the best results for each model and task in bold.}
\label{tab:main_results_direct_self}
\end{table*}

\subsection{Self-translate vs. MT}

We include additional self-translate vs. MT results for \textbf{XGLM} \citep{lin-etal-2022-shot} and \textbf{LLaMA} \cite{touvron2023llama}. \cref{tab:main_results_mt} shows task accuracy for different sizes of these models, using \textbf{self-translate} inference and \textbf{MT}. The last column shows the average accuracy over all tasks.

\begin{table*}[t]
\centering
\small
\begin{adjustbox}{max width=\linewidth}
\begin{tabular}{crllllll|l}
\toprule
Model & Size & Method & \multicolumn{1}{c}{XStoryC} & \multicolumn{1}{c}{XCOPA} & \multicolumn{1}{c}{XNLI} & \multicolumn{1}{c}{PAWS-X} & \multicolumn{1}{c}{MGSM} & \multicolumn{1}{|c}{Avg} \\ \midrule
\multirow{9.5}{*}{XGLM}
& \multirow{2}{*}{0.6B}
& Self-translate &          52.8 &   53.4 &  41.5 &   50.6  & \textbf{1.4} & 39.9 \\
  &&       MT &          \textbf{57.3} &   \textbf{59.8} &  \textbf{46.3} &   \textbf{51.7} &   1.1 & \textbf{43.2} \\
\cmidrule{2-9}
& \multirow{2}{*}{1.7B}
& Self-translate &          55.9 &   58.4 &  44.9 &   50.2  &   1.7 & 42.2 \\
  &&       MT &          \textbf{60.7} &   \textbf{62.3} &  \textbf{47.4} &   \textbf{51.2}  &   \textbf{2.3} & \textbf{44.8} \\
\cmidrule{2-9}
& \multirow{2}{*}{2.9B}
& Self-translate &          58.2 &   62.5 &  46.2 &   53.2  &   1.6 & 44.3\\
  &&       MT &          \textbf{62.3} &   \textbf{65.3} &  \textbf{48.8} &   \textbf{55.7}  &   \textbf{2.2} & \textbf{46.9} \\
\cmidrule{2-9}
& \multirow{2}{*}{7.5B}
& Self-translate &          60.9 &   64.4 &  48.9 &   55.4  &   \textbf{0.1} & 45.9 \\
  &&       MT &          \textbf{63.6} &   \textbf{66.3} &  \textbf{50.7} &   \textbf{57.4} &   0.0  & \textbf{47.6}\\

\midrule

\multirow{7}{*}{LLaMA}
& \multirow{2}{*}{7B}
& Self-translate &          55.8 &   54.9 &  43.0 &   57.0 &   6.1 & 43.4 \\
   &&       MT &          \textbf{66.8} &   \textbf{68.6} &  \textbf{48.6 }&   \textbf{58.8} &  \textbf{10.7} & \textbf{50.7} \\
\cmidrule{2-9}
& \multirow{2}{*}{13B}
& Self-translate &          57.7 &   56.5 &  \textbf{35.1} &   52.1 &  10.0 & 42.3 \\
  &&       MT &          \textbf{68.1} &   \textbf{70.4} &  \textbf{35.1} &   \textbf{54.2} &  \textbf{16.5} & \textbf{48.9} \\
\cmidrule{2-9}
& \multirow{2}{*}{30B}
& Self-translate &          59.0 &   58.4 &  43.5 &   55.6 &  16.3 & 46.6 \\
  &&       MT &          \textbf{68.7} &   \textbf{71.5} &  \textbf{46.1} &   \textbf{55.9} &  \textbf{28.6} & \textbf{54.2} \\
\bottomrule
\end{tabular}
\end{adjustbox}
\caption{\textbf{Self-translate vs. MT.} Task accuracy for different sizes of XGLM and LLaMA, using self-translate and MT. The last column shows the average accuracy over all tasks. We highlight the best results for each model and task in bold.}
\label{tab:main_results_mt}
\end{table*}

\subsection{Results by language}
\label{sec:tables_by_language}

We include additional language results for \textbf{XGLM} \citep{lin-etal-2022-shot} and \textbf{LLaMA} \cite{touvron2023llama}. \cref{tab:xstory_cloze,tab:xcopa,tab:xnli,tab:pawsx,tab:mgsm} show the results by language in different tasks, using different model sizes and the \textbf{direct} inference, \textbf{self-translate}, and \textbf{MT} methods. The last column shows the average accuracy over all languages except English.

\begin{table*}[ht]
\centering
\small
\begin{adjustbox}{max width=\textwidth}
\begin{tabular}{lrlrrrrrrrrrrr|r}
\toprule
  Model & Size & Method  &   ar &   en &   es &   eu &   hi &   id &   my &   ru &   sw &   te &   zh &  avg \\
\midrule
\multirow{13.5}{*}{XGLM}
 & \multirow{3}{*}{0.6B} & Direct & 50.1 & 60.6 & 55.1 & 53.1 & 52.3 & 54.0 & 51.5 & 56.2 & 53.1 & 55.9 & 53.3 & 53.5 \\
 &  &  Self-translate & 52.2 &    \_ & 53.1 & 54.0 & 53.5 & 53.6 & 52.3 & 53.9 & 52.1 & 53.0 & 50.0 & 52.8 \\
 &  &       MT & 58.1 &    \_ & 57.2 & 55.7 & 57.4 & 57.9 & 55.2 & 58.8 & 56.5 & 59.5 & 56.8 & 57.3 \\ \cmidrule{2-15}
 & \multirow{3}{*}{1.7B} & Direct & 52.5 & 64.3 & 59.2 & 56.1 & 55.8 & 58.0 & 53.8 & 59.8 & 56.0 & 58.0 & 56.2 & 56.5 \\
 &  &  Self-translate & 55.4 &    \_ & 58.4 & 54.3 & 55.1 & 57.1 & 55.5 & 58.4 & 55.3 & 54.8 & 54.9 & 55.9 \\
 &  &       MT & 61.9 &    \_ & 60.4 & 58.3 & 61.7 & 61.4 & 57.8 & 62.7 & 60.0 & 61.3 & 61.6 & 60.7 \\ \cmidrule{2-15}
 &  \multirow{3}{*}{2.9B} & Direct & 53.9 & 67.3 & 61.0 & 56.3 & 57.5 & 61.4 & 55.2 & 62.2 & 56.7 & 60.0 & 57.6 & 58.2 \\
 &  &  Self-translate & 56.3 &    \_ & 61.3 & 56.9 & 58.3 & 60.4 & 57.6 & 59.7 & 57.9 & 56.3 & 57.8 & 58.2 \\
 &  &       MT & 63.0 &    \_ & 63.2 & 61.2 & 63.3 & 62.9 & 58.8 & 64.7 & 60.0 & 62.8 & 63.0 & 62.3 \\ \cmidrule{2-15}
 &  \multirow{3}{*}{7.5B} & Direct & 56.2 & 69.8 & 64.1 & 57.7 & 58.8 & 62.9 & 57.1 & 63.5 & 59.3 & 60.2 & 58.9 & 59.9 \\
 &  &  Self-translate & 60.7 &    \_ & 63.8 & 59.8 & 61.3 & 62.9 & 57.8 & 64.4 & 60.0 & 57.6 & 60.4 & 60.9 \\
 &  &       MT & 64.3 &    \_ & 64.7 & 63.1 & 64.9 & 63.4 & 60.3 & 65.9 & 61.4 & 63.3 & 65.0 & 63.6 \\
  \midrule
  \multirow{10}{*}{LLaMA}
   &  \multirow{3}{*}{7B} & Direct & 48.3 & 74.8 & 65.1 & 50.1 & 52.7 & 52.1 & 48.7 & 61.4 & 50.4 & 52.9 & 54.3 & 53.6 \\
   & &  Self-translate & 52.2 &    \_ & 68.0 & 50.0 & 51.9 & 56.5 & 50.2 & 66.8 & 50.6 & 51.4 & 60.4 & 55.8 \\
   & &       MT & 67.7 &    \_ & 68.4 & 65.4 & 68.5 & 68.3 & 62.5 & 70.1 & 64.3 & 65.5 & 67.2 & 66.8 \\\cmidrule{2-15}
 & \multirow{3}{*}{13B} & Direct & 49.7 & 77.3 & 69.4 & 50.7 & 52.3 & 55.3 & 47.8 & 63.4 & 49.9 & 53.3 & 56.5 & 54.8 \\
  & &  Self-translate & 55.2 &    \_ & 72.1 & 50.8 & 53.7 & 59.3 & 51.8 & 70.4 & 48.4 & 51.8 & 63.2 & 57.7 \\
  & &       MT & 68.6 &    \_ & 70.0 & 66.4 & 70.0 & 69.0 & 62.8 & 71.7 & 66.0 & 67.7 & 69.1 & 68.1 \\\cmidrule{2-15}
 & \multirow{3}{*}{30B} & Direct & 50.9 & 78.2 & 70.8 & 51.4 & 56.7 & 59.2 & 48.8 & 66.7 & 50.6 & 53.2 & 58.6 & 56.7 \\
  & &  Self-translate & 56.4 &    \_ & 74.0 & 48.8 & 60.2 & 62.6 & 51.0 & 71.4 & 48.9 & 49.9 & 67.0 & 59.0 \\
 & &       MT & 70.0 &    \_ & 71.5 & 66.6 & 70.0 & 69.3 & 63.6 & 73.3 & 67.0 & 66.9 & 69.0 & 68.7 \\
\bottomrule
\end{tabular}
\end{adjustbox}
\caption{\textbf{XGLM and LLaMA results on XStoryCloze for each language.} We show task accuracy for different sizes of these models, using \textbf{direct} inference \textbf{self-translate} and \textbf{MT}. The last column shows the average accuracy over all languages except English.}
\label{tab:xstory_cloze}
\end{table*}

\begin{table*}[ht]
\centering
\small
\begin{adjustbox}{max width=\textwidth}
\begin{tabular}{lrlrrrrrrrrrrr|r}
\toprule
  Model & Size & Method  &   et &   ht &   id &   it &   qu &   sw &   ta &   th &   tr &   vi &   zh &  avg \\
\midrule
\multirow{13.5}{*}{XGLM}
 & \multirow{3}{*}{0.6B} & Direct & 55.6 & 55.0 & 57.2 & 53.8 & 49.2 & 53.2 & 56.2 & 55.2 & 54.4 & 58.4 & 55.6 & 54.9 \\
 &  &  Self-translate & 52.2 & 54.2 & 59.4 & 51.8 & 50.0 & 52.6 & 55.0 & 55.2 & 55.2 & 51.8 & 50.4 & 53.4 \\
 &  &       MT & 60.0 & 61.0 & 60.4 & 61.8 & 50.4 & 59.4 & 61.6 & 58.8 & 62.4 & 61.8 & 60.2 & 59.8 \\\cmidrule{2-15}
 & \multirow{3}{*}{1.7B} & Direct & 56.8 & 55.8 & 64.6 & 54.0 & 52.2 & 56.6 & 55.2 & 58.2 & 53.4 & 63.0 & 58.0 & 57.1 \\
 &  &  Self-translate & 59.0 & 57.0 & 60.6 & 60.0 & 50.8 & 57.8 & 58.8 & 58.4 & 60.8 & 61.0 & 58.4 & 58.4 \\
 &  &       MT & 65.6 & 62.8 & 63.4 & 65.6 & 50.4 & 62.2 & 63.8 & 61.0 & 63.8 & 64.0 & 62.6 & 62.3 \\\cmidrule{2-15}
 & \multirow{3}{*}{2.9B} & Direct & 58.2 & 55.8 & 66.8 & 60.2 & 50.2 & 58.8 & 54.2 & 57.0 & 56.6 & 65.2 & 60.0 & 58.5 \\
 &  &  Self-translate & 64.4 & 65.2 & 64.8 & 64.2 & 52.0 & 62.2 & 59.4 & 60.8 & 62.0 & 65.4 & 67.4 & 62.5 \\
 &  &       MT & 69.2 & 65.4 & 67.2 & 70.8 & 51.0 & 64.8 & 65.2 & 64.0 & 66.4 & 67.2 & 67.0 & 65.3 \\\cmidrule{2-15}
 & \multirow{3}{*}{7.5B} & Direct & 61.2 & 57.4 & 69.4 & 63.6 & 48.8 & 60.0 & 54.4 & 59.4 & 58.4 & 70.2 & 63.8 & 60.6 \\
 &  &  Self-translate & 66.8 & 64.6 & 66.8 & 68.4 & 51.0 & 62.8 & 65.6 & 62.8 & 65.4 & 65.2 & 68.6 & 64.4 \\
 &  &       MT & 71.8 & 64.8 & 67.6 & 72.8 & 50.4 & 66.8 & 67.4 & 62.0 & 69.8 & 68.6 & 67.6 & 66.3 \\
  \midrule
  \multirow{10}{*}{LLaMA}
 &  \multirow{3}{*}{7B} & Direct & 48.8 & 51.0 & 54.6 & 62.0 & 51.4 & 50.8 & 55.2 & 55.8 & 55.6 & 51.6 & 56.2 & 53.9 \\
  &  &  Self-translate & 54.2 & 51.2 & 59.4 & 73.8 & 48.4 & 52.8 & 47.6 & 50.8 & 51.6 & 47.8 & 66.0 & 54.9 \\
  &  &       MT & 72.6 & 68.2 & 71.0 & 75.4 & 52.2 & 67.4 & 70.2 & 62.2 & 72.6 & 71.2 & 71.6 & 68.6 \\\cmidrule{2-15}
 & \multirow{3}{*}{13B} & Direct & 48.2 & 52.8 & 57.8 & 67.2 & 50.2 & 51.2 & 54.4 & 54.6 & 53.0 & 53.8 & 58.4 & 54.7 \\
 & &  Self-translate & 51.8 & 51.4 & 62.8 & 75.8 & 51.6 & 49.4 & 51.2 & 51.4 & 56.6 & 49.2 & 69.8 & 56.5 \\
  & &       MT & 73.2 & 70.0 & 72.8 & 76.8 & 51.6 & 70.2 & 71.8 & 64.8 & 73.2 & 75.2 & 75.2 & 70.4 \\\cmidrule{2-15}
 & \multirow{3}{*}{30B} & Direct & 47.2 & 51.8 & 60.6 & 71.4 & 49.4 & 52.4 & 53.2 & 54.6 & 52.2 & 52.4 & 62.2 & 55.2 \\
 &  &  Self-translate & 50.4 & 53.0 & 68.0 & 79.0 & 49.4 & 50.2 & 52.8 & 48.6 & 59.8 & 58.4 & 73.2 & 58.4 \\
 &  &       MT & 75.2 & 71.2 & 73.2 & 80.6 & 52.6 & 70.6 & 72.2 & 64.6 & 74.2 & 75.0 & 76.8 & 71.5 \\
\bottomrule
\end{tabular}
\end{adjustbox}
\caption{\textbf{XGLM and LLaMA results on XCOPA for each language.} We show task accuracy for different sizes of these models, using \textbf{direct} inference \textbf{self-translate} and \textbf{MT}. The last column shows the average accuracy over all languages.}
\label{tab:xcopa}
\end{table*}

\begin{table*}[ht]
\centering
\begin{adjustbox}{max width=\textwidth}
\begin{tabular}{lrlrrrrrrrrrrrrrrr|r}
\toprule
 Model & Size & Method  &   ar &   bg &   de &   el &   en &   es &   fr &   hi &   ru &   sw &   th &   tr &   ur &   vi &   zh &  avg \\
\midrule
\multirow{13.5}{*}{XGLM}
& \multirow{3}{*}{0.6B} & Direct & 33.4 & 41.3 & 44.5 & 39.6 & 48.3 & 42.0 & 45.5 & 38.7 & 44.6 & 36.1 & 38.8 & 40.2 & 34.5 & 38.5 & 33.5 & 39.4 \\
 &  &  Self-translate & 40.2 & 43.9 & 43.9 & 42.2 &    \_ & 43.3 & 43.3 & 41.4 & 43.0 & 39.0 & 41.9 & 40.6 & 40.6 & 41.5 & 35.8 & 41.5 \\
 &  &       MT & 46.9 & 47.1 & 46.6 & 46.6 &    \_ & 47.5 & 46.5 & 45.6 & 45.7 & 45.6 & 46.3 & 46.4 & 43.8 & 46.8 & 47.1 & 46.3 \\\cmidrule{2-19}
 & \multirow{3}{*}{1.7B} & Direct & 33.5 & 44.7 & 45.3 & 40.1 & 49.7 & 43.6 & 45.7 & 42.6 & 46.0 & 42.0 & 41.7 & 43.0 & 39.5 & 45.0 & 33.8 & 41.9 \\
 &  &  Self-translate & 44.2 & 46.8 & 47.0 & 46.1 &    \_ & 45.9 & 46.8 & 44.1 & 45.7 & 43.8 & 44.0 & 42.7 & 42.0 & 44.7 & 44.3 & 44.9 \\
 & &       MT & 47.3 & 47.8 & 48.8 & 48.1 &    \_ & 48.5 & 48.6 & 47.1 & 47.2 & 45.9 & 46.5 & 48.3 & 44.2 & 48.6 & 47.3 & 47.4 \\\cmidrule{2-19}
 & \multirow{3}{*}{2.9B} & Direct & 33.7 & 46.0 & 48.3 & 41.4 & 51.1 & 46.7 & 45.0 & 44.0 & 45.3 & 44.4 & 42.0 & 45.0 & 40.1 & 46.0 & 34.8 & 43.0 \\
 &  &  Self-translate & 43.9 & 48.1 & 48.4 & 47.3 &    \_ & 48.2 & 48.5 & 44.1 & 46.5 & 44.8 & 45.8 & 45.2 & 42.4 & 46.6 & 46.7 & 46.2 \\
 &  &       MT & 48.9 & 49.5 & 50.0 & 49.4 &    \_ & 50.5 & 50.0 & 48.5 & 47.9 & 47.7 & 47.5 & 48.6 & 45.4 & 49.6 & 49.0 & 48.8 \\\cmidrule{2-19}
 & \multirow{3}{*}{7.5B} & Direct & 33.4 & 44.9 & 49.0 & 40.7 & 53.9 & 47.7 & 46.9 & 47.2 & 46.3 & 45.8 & 43.7 & 46.3 & 42.1 & 46.3 & 35.4 & 44.0 \\
 &  &  Self-translate & 47.0 & 51.6 & 50.4 & 50.7 &    \_ & 51.8 & 51.6 & 46.8 & 50.0 & 47.3 & 47.4 & 47.5 & 44.5 & 48.9 & 48.6 & 48.9 \\
 &  &       MT & 50.6 & 51.8 & 51.8 & 51.6 &    \_ & 52.8 & 52.1 & 51.0 & 50.5 & 48.7 & 48.6 & 51.8 & 46.9 & 50.2 & 51.2 & 50.7 \\
  \midrule
  \multirow{10}{*}{LLaMA}
 &  \multirow{3}{*}{7B} & Direct & 33.6 & 37.0 & 44.8 & 34.9 & 51.1 & 40.6 & 43.8 & 36.1 & 39.4 & 33.7 & 34.5 & 35.6 & 33.4 & 35.6 & 36.2 & 37.1 \\
 &   &  Self-translate & 40.7 & 48.7 & 50.6 & 43.5 &    \_ & 49.8 & 49.5 & 39.7 & 48.0 & 34.8 & 36.3 & 38.0 & 36.4 & 39.9 & 46.1 & 43.0 \\
  &  &       MT & 48.6 & 49.3 & 49.9 & 50.1 &    \_ & 50.4 & 50.1 & 48.5 & 48.3 & 46.5 & 46.4 & 48.0 & 45.5 & 49.2 & 49.3 & 48.6 \\\cmidrule{2-19}
 & \multirow{3}{*}{13B} & Direct & 34.1 & 34.1 & 35.3 & 34.8 & 35.7 & 33.4 & 33.4 & 35.5 & 34.1 & 33.0 & 34.5 & 34.0 & 34.3 & 34.0 & 34.4 & 34.2 \\
  & &  Self-translate & 35.3 & 34.7 & 35.3 & 35.1 &    \_ & 36.0 & 35.8 & 35.4 & 35.0 & 34.9 & 34.8 & 34.6 & 34.9 & 35.4 & 34.4 & 35.1 \\
  & &       MT & 34.1 & 35.3 & 35.3 & 35.5 &    \_ & 35.2 & 35.2 & 35.3 & 35.3 & 35.2 & 34.1 & 34.6 & 35.0 & 34.8 & 36.1 & 35.1 \\\cmidrule{2-19}
 & \multirow{3}{*}{30B} & Direct & 34.4 & 38.6 & 44.0 & 35.1 & 47.9 & 40.4 & 42.9 & 36.6 & 38.2 & 34.2 & 34.0 & 36.3 & 34.3 & 35.6 & 33.6 & 37.0 \\
 &  &  Self-translate & 42.2 & 47.6 & 47.7 & 44.8 &    \_ & 48.1 & 47.8 & 41.4 & 47.3 & 37.3 & 37.4 & 42.0 & 38.9 & 41.6 & 44.3 & 43.5 \\
 &  &       MT & 46.2 & 46.4 & 47.3 & 46.9 &    \_ & 47.7 & 47.4 & 45.7 & 46.3 & 44.8 & 45.0 & 45.3 & 43.8 & 46.5 & 46.6 & 46.1 \\
\bottomrule
\end{tabular}
\end{adjustbox}
\caption{\textbf{XGLM and LLaMA results on XNLI for each language.} We show task accuracy for different sizes of these models, using \textbf{direct} inference \textbf{self-translate} and \textbf{MT}. The last column shows the average accuracy over all languages except English.}
\label{tab:xnli}
\end{table*}

\begin{table*}[ht]
\centering
\small
\begin{adjustbox}{max width=\textwidth}
\begin{tabular}{lrlrrrrrrr|r}
\toprule
 Model & Size & Method  & de &   en &   es &   fr &   ja &   ko &   zh &  avg \\
\midrule
\multirow{13.5}{*}{XGLM}
 & \multirow{3}{*}{0.6B} & Direct & 49.1 & 50.6 & 52.5 & 50.8 & 44.1 & 46.2 & 47.8 & 48.4 \\
  & &  Self-translate & 51.1 &    \_ & 50.1 & 50.3 & 50.9 & 50.4 & 51.0 & 50.6 \\
 &  &       MT & 53.5 &    \_ & 52.8 & 51.0 & 51.2 & 50.4 & 51.2 & 51.7 \\\cmidrule{2-11}
 & \multirow{3}{*}{1.7B} & Direct & 57.6 & 52.6 & 53.8 & 47.3 & 46.1 & 51.4 & 48.1 & 50.7 \\
  & &  Self-translate & 50.0 &    \_ & 51.6 & 51.6 & 49.6 & 49.1 & 49.4 & 50.2 \\
  & &       MT & 51.9 &    \_ & 51.6 & 52.8 & 50.2 & 51.1 & 49.5 & 51.2 \\\cmidrule{2-11}
 & \multirow{3}{*}{2.9B} & Direct & 50.6 & 54.8 & 53.1 & 49.7 & 50.9 & 46.8 & 53.7 & 50.8 \\
  & &  Self-translate & 54.9 &    \_ & 53.9 & 54.2 & 52.1 & 51.6 & 52.7 & 53.2 \\
  & &       MT & 56.5 &    \_ & 57.0 & 56.2 & 54.8 & 54.5 & 55.4 & 55.7 \\\cmidrule{2-11}
 & \multirow{3}{*}{7.5B} & Direct & 55.9 & 58.9 & 52.8 & 51.8 & 52.0 & 46.0 & 51.3 & 51.6 \\
  & &  Self-translate & 57.7 &    \_ & 56.1 & 56.1 & 54.5 & 53.0 & 54.9 & 55.4 \\
  & &       MT & 59.6 &    \_ & 58.4 & 59.0 & 54.6 & 55.2 & 57.7 & 57.4 \\
  \midrule
  \multirow{10}{*}{LLaMA}
  & \multirow{3}{*}{7B} & Direct & 54.6 & 61.9 & 56.1 & 52.9 & 56.7 & 49.7 & 49.1 & 53.2 \\
  &  &  Self-translate & 59.8 &    \_ & 60.7 & 59.2 & 53.9 & 52.5 & 55.8 & 57.0 \\
  &  &       MT & 59.9 &    \_ & 60.6 & 60.1 & 57.6 & 57.5 & 57.3 & 58.8 \\\cmidrule{2-11}
 & \multirow{3}{*}{13B} & Direct & 52.9 & 53.1 & 52.4 & 54.6 & 45.0 & 46.9 & 45.2 & 49.5 \\
  & &  Self-translate & 52.9 &    \_ & 52.5 & 52.9 & 51.2 & 51.6 & 51.5 & 52.1 \\
  & &       MT & 53.6 &    \_ & 54.4 & 53.8 & 55.3 & 54.4 & 53.8 & 54.2 \\\cmidrule{2-11}
 & \multirow{3}{*}{30B} & Direct & 58.4 & 58.5 & 56.0 & 52.5 & 46.6 & 45.6 & 46.2 & 50.9 \\
  & &  Self-translate & 56.5 &    \_ & 56.8 & 58.1 & 54.5 & 52.1 & 55.5 & 55.6 \\
  & &       MT & 56.6 &    \_ & 57.8 & 56.9 & 55.1 & 54.8 & 54.2 & 55.9 \\
\bottomrule
\end{tabular}
\end{adjustbox}
\caption{\textbf{XGLM and LLaMA results on PAWS-X for each language.} We show task accuracy for different sizes of these models, using \textbf{direct} inference \textbf{self-translate} and \textbf{MT}. The last column shows the average accuracy over all languages except English.}
\label{tab:pawsx}
\end{table*}

\begin{table*}[ht]
\centering
\small
\begin{adjustbox}{max width=\textwidth}
\begin{tabular}{lrlrrrrrrrrrrr|r}
\toprule
 Model & Size & Method &   bn &   de &   en &   es &   fr &   ja &   ru &   sw &   te &   th &   zh &  avg \\
\midrule
\multirow{13.5}{*}{XGLM}
  &   \multirow{3}{*}{0.6B} &  Direct & 1.2 & 0.8 & 2.0 & 1.2 & 1.6 & 4.0 & 0.4 & 2.4 & 0.4 & 1.6 & 3.2 &  1.7 \\
  &    & Self-translate & 0.0 & 2.0 &   \_ & 2.0 & 1.6 & 0.8 & 1.2 & 2.0 & 2.4 & 0.8 & 1.6 &  1.4 \\
  &   &      MT & 1.2 & 1.2 &   \_ & 0.8 & 0.8 & 2.0 & 1.6 & 1.2 & 0.4 & 1.6 & 0.0 &  1.1 \\ \cmidrule{2-15}
  &   \multirow{3}{*}{1.7B} &  Direct & 0.8 & 1.2 & 2.0 & 2.4 & 2.0 & 1.6 & 0.8 & 1.2 & 2.0 & 2.0 & 2.8 &  1.7 \\
  &    & Self-translate & 1.2 & 2.0 &   \_ & 2.8 & 1.6 & 2.4 & 2.8 & 1.2 & 1.2 & 0.8 & 1.2 &  1.7 \\
  &    &      MT & 2.0 & 2.4 &   \_ & 2.0 & 0.8 & 2.8 & 2.0 & 2.8 & 3.2 & 2.8 & 2.4 &  2.3 \\ \cmidrule{2-15}
  &   \multirow{3}{*}{2.9B} &  Direct & 0.0 & 0.8 & 2.4 & 2.0 & 1.2 & 2.0 & 2.0 & 2.0 & 2.0 & 0.8 & 1.2 &  1.4 \\
  &    & Self-translate & 0.8 & 1.2 &   \_ & 1.6 & 1.6 & 1.6 & 1.2 & 2.0 & 1.2 & 2.4 & 2.0 &  1.6 \\
  &    &      MT & 2.8 & 2.4 &   \_ & 2.8 & 2.4 & 1.2 & 1.6 & 2.0 & 3.2 & 0.8 & 2.4 &  2.2 \\ \cmidrule{2-15}
  &   \multirow{3}{*}{7.5B} &  Direct & 0.0 & 1.2 & 0.0 & 0.0 & 0.0 & 0.4 & 2.4 & 0.4 & 1.2 & 1.6 & 1.2 &  0.8 \\
  &    & Self-translate & 0.0 & 0.4 &   \_ & 0.0 & 0.0 & 0.0 & 0.4 & 0.0 & 0.4 & 0.0 & 0.0 &  0.1 \\
  &    &      MT & 0.0 & 0.0 &   \_ & 0.0 & 0.0 & 0.0 & 0.0 & 0.0 & 0.0 & 0.4 & 0.0 &  0.0 \\
 \midrule
 \multirow{10}{*}{LLaMA}
   & \multirow{3}{*}{7B} & Direct &  0.0 &  9.6 & 13.6 & 10.4 &  8.8 &  5.2 & 10.0 &  2.0 &  0.0 &  0.0 &  4.4 &  5.0 \\
   &  &  Self-translate &  2.0 & 11.2 &    \_ & 11.2 & 12.4 &  4.8 & 10.8 &  1.2 &  0.4 &  2.4 &  4.8 &  6.1 \\
   &  &       MT & 10.0 & 12.4 &    \_ & 12.0 &  9.6 & 10.8 & 10.8 & 12.0 &  9.6 &  8.4 & 11.2 & 10.7 \\\cmidrule{2-15}
  & \multirow{3}{*}{13B} & Direct &  0.0 & 16.0 & 20.8 & 15.2 & 15.6 &  5.2 & 10.0 &  3.6 &  0.0 &  0.0 &  8.8 &  7.4 \\
  &  &  Self-translate &  3.6 & 17.6 &    \_ & 20.4 & 18.0 &  9.2 & 15.2 &  3.6 &  0.0 &  1.6 & 10.4 & 10.0 \\
  &  &       MT & 16.8 & 20.0 &    \_ & 20.8 & 15.2 & 15.2 & 15.6 & 19.2 & 14.0 & 14.0 & 14.4 & 16.5 \\\cmidrule{2-15}
  & \multirow{3}{*}{30B} & Direct &  0.0 & 29.2 & 39.6 & 33.2 & 30.4 &  7.2 & 27.2 &  5.2 &  0.0 &  0.0 & 22.8 & 15.5 \\
  &  &  Self-translate &  8.0 & 34.4 &    \_ &  9.6 & 24.4 & 20.8 & 29.6 &  6.4 &  0.4 &  3.6 & 25.6 & 16.3 \\
  & &       MT & 28.4 & 32.4 &    \_ & 31.2 & 35.2 & 29.2 & 26.4 & 32.0 & 25.6 & 20.0 & 25.6 & 28.6 \\
\bottomrule
\end{tabular}
\end{adjustbox}
\caption{\textbf{XGLM and LLaMA results on MGSM for each language.} We show task accuracy for different sizes of these models, using \textbf{direct} inference \textbf{self-translate} and \textbf{MT}. The last column shows the average accuracy over all languages except English.}
\label{tab:mgsm}
\end{table*}

\subsection{Translation metrics}
\label{app:metrics}

We obtain similar results with BLEU \cite{papineni2002bleu} and COMET \cite{rei2022comet} metrics. We report the average COMET and BLEU scores across all languages for NLLB, XGLM, BLOOM and LLaMA in \cref{tab:translation_metrics_comet,tab:translation_metrics}. 

\begin{table*}[ht]
\centering
\small
\begin{adjustbox}{max width=\textwidth}
\begin{tabular}{lrrrrrrr}
\toprule
Model &  Size &  XStoryC &  XCOPA &  XNLI &  PAWS-X &  MGSM &  Avg \\
\midrule
 \multirow{4}{*}{NLLB} &   0.6B &          86.9 &   80.3 &  84.6 &    85.4 &  80.2 & 83.5 \\
  &   1.3B &          88.2 &   82.9 &  85.6 &    86.0 &  83.8 & 85.3 \\
  &   1.3B &          88.3 &   82.1 &  85.5 &    86.0 &  83.5 & 85.1 \\
  &   3.3B &          88.7 &   83.3 &  85.9 &    86.2 &  84.5 & 85.7 \\
 \midrule
 \multirow{4}{*}{XGLM} &   0.6B &          63.4 &   61.3 &  66.2 &    66.0 &  54.7 & 62.3 \\
  &   1.7B &          77.1 &   74.1 &  75.8 &    75.9 &  68.4 & 74.3 \\
  &   2.9B &          81.1 &   77.6 &  78.5 &    79.2 &  73.5 & 78.0 \\
  &   7.5B &          84.2 &   79.8 &  81.7 &    81.6 &  79.2 & 81.3 \\
 \midrule
\multirow{4}{*}{BLOOM} &   0.6B &          61.5 &   54.0 &  63.6 &    60.6 &  48.2 & 57.6 \\
 &   1.7B &          73.6 &   61.9 &  67.4 &    72.1 &  61.7 & 67.3 \\
 &   3B &          76.3 &   63.3 &  69.5 &    74.7 &  69.1 & 70.6 \\
 &   7.1B &          78.8 &   66.4 &  73.1 &    78.8 &  74.5 & 74.3 \\
\midrule
\multirow{3}{*}{LLaMA} &   7B &          66.8 &   59.4 &  71.5 &    80.9 &  66.0 & 68.9 \\
 &  13B &          68.8 &   61.8 &  75.0 &    82.6 &  69.6 & 71.6 \\
 &  30B &          71.7 &   65.0 &  78.4 &    83.8 &  67.5 & 73.3 \\
\bottomrule
\end{tabular}
\end{adjustbox}
\caption{COMET translation metrics for different models.}
\label{tab:translation_metrics_comet}
\end{table*}

\begin{table*}[ht]
\centering
\small
\begin{adjustbox}{max width=\textwidth}
\begin{tabular}{lrrrrrrr}
\toprule
Model &  Size &  XStoryC &  XCOPA &  XNLI &  PAWS-X &  MGSM &  Avg \\
\midrule
 \multirow{4}{*}{NLLB} &   0.6B &          38.0 &   32.1 &  38.0 &    49.0 &  32.1 & 37.8 \\
  &   1.3B &          40.6 &   36.6 &  40.3 &    51.3 &  41.3 & 42.0 \\
  &   1.3B &          40.9 &   35.6 &  40.1 &    50.9 &  40.9 & 41.7 \\
  &   3.3B &          41.8 &   37.6 &  41.5 &    51.9 &  43.7 & 43.3 \\
 \midrule
 \multirow{4}{*}{XGLM} &   0.6B &           7.1 &    6.5 &  10.4 &    18.0 &   5.4 &  9.5 \\
  &   1.7B &          18.5 &   18.1 &  20.3 &    28.3 &  17.1 & 20.5 \\
  &   2.9B &          23.8 &   24.1 &  24.1 &    33.1 &  23.5 & 25.7 \\
  &   7.5B &          29.0 &   28.4 &  28.8 &    37.0 &  28.3 & 30.3 \\
 \midrule
\multirow{4}{*}{BLOOM} &   0.6B &           7.9 &    4.8 &  11.8 &    16.2 &   5.4 &  9.2 \\
 &   1.7B &          17.3 &   10.5 &  14.9 &    27.2 &  12.6 & 16.5 \\
 &   3B &          20.2 &   13.0 &  17.1 &    31.1 &  20.3 & 20.3 \\
 &   7.1B &          25.2 &   16.5 &  21.4 &    36.1 &  27.7 & 25.4 \\
\midrule
\multirow{3}{*}{LLaMA} &   7B &          14.7 &    8.9 &  19.9 &    39.1 &  23.9 & 21.3 \\
 &  13B &          17.7 &   12.4 &  24.1 &    42.5 &  27.9 & 24.9 \\
 &  30B &          21.2 &   15.4 &  27.7 &    45.4 &  25.5 & 27.0 \\
\bottomrule
\end{tabular}
\end{adjustbox}
\caption{BLEU translation metrics for different models.}
\label{tab:translation_metrics}
\end{table*}

\subsection{Translation metrics by language}
\label{app:metrics_language}

We report NLLB, XGLM, BLOOM and LLaMA COMET metrics for each language and task in \cref{tab:xstory_cloze_comet,tab:xcopa_comet,tab:xnli_comet,tab:pawsx_comet,tab:mgsm_comet}, and BLEU metrics in \cref{tab:xstory_cloze_metrics,tab:xcopa_metrics,tab:xnli_metrics,tab:pawsx_metrics,tab:mgsm_metrics}.

\begin{table*}[ht]
\centering
\small
\begin{adjustbox}{max width=\textwidth}
\begin{tabular}{lrrrrrrrrrrrr}
\toprule
Model &  Size &    ru &    zh &    es &    ar &    hi &    id &    te &    sw &    eu &    my &  avg \\
\midrule
 \multirow{4}{*}{NLLB} &   0.6B & 87.07 & 85.00 & 89.36 & 88.39 & 90.52 & 88.08 & 86.44 & 86.04 & 86.87 & 81.35 & 86.9 \\
  &   1.3B & 88.44 & 86.02 & 90.33 & 89.85 & 91.56 & 89.14 & 87.64 & 87.31 & 86.92 & 85.26 & 88.2 \\
  &   1.3B & 88.18 & 86.36 & 90.22 & 89.83 & 91.39 & 89.05 & 87.30 & 87.21 & 87.25 & 85.99 & 88.3 \\
  &   3.3B & 88.63 & 87.54 & 90.54 & 90.36 & 91.70 & 89.54 & 88.00 & 87.46 & 86.92 & 86.60 & 88.7 \\
 \midrule
 \multirow{4}{*}{XGLM} &   0.6B & 73.05 & 54.47 & 72.08 & 61.44 & 68.85 & 77.52 & 57.04 & 58.63 & 59.52 & 50.99 & 63.4 \\
  &   1.7B & 80.96 & 77.26 & 81.95 & 76.35 & 77.48 & 83.96 & 74.09 & 75.15 & 71.25 & 73.03 & 77.1 \\
  &   2.9B & 83.36 & 82.11 & 85.61 & 79.84 & 82.99 & 85.66 & 75.43 & 79.71 & 79.32 & 77.47 & 81.1 \\
  &   7.5B & 85.76 & 84.25 & 87.81 & 83.81 & 86.25 & 87.60 & 80.66 & 82.92 & 82.05 & 81.36 & 84.2 \\
 \midrule
\multirow{4}{*}{BLOOM} &   0.6B & 43.20 & 70.47 & 73.65 & 72.18 & 73.40 & 79.31 & 58.06 & 42.03 & 55.73 & 47.25 & 61.5 \\
 &   1.7B & 60.47 & 82.81 & 85.44 & 80.40 & 81.05 & 85.06 & 72.48 & 66.06 & 71.98 & 50.69 & 73.6 \\
 &   3B & 63.44 & 84.45 & 87.16 & 82.20 & 83.16 & 85.72 & 75.11 & 71.03 & 76.99 & 53.68 & 76.3 \\
 &   7.1B & 68.97 & 86.63 & 88.42 & 84.68 & 86.76 & 87.87 & 78.86 & 75.15 & 80.88 & 49.80 & 78.8 \\
\midrule
\multirow{3}{*}{LLaMA} &   7B & 85.66 & 79.10 & 88.56 & 65.12 & 67.96 & 77.08 & 50.39 & 52.14 & 49.66 & 52.55 & 66.8 \\
 &  13B & 87.02 & 82.66 & 89.37 & 70.64 & 72.86 & 81.15 & 48.62 & 53.14 & 51.36 & 51.17 & 68.8 \\
 &  30B & 87.98 & 84.37 & 90.13 & 77.37 & 81.64 & 84.55 & 49.38 & 59.99 & 52.50 & 49.04 & 71.7 \\
\bottomrule
\end{tabular}
\end{adjustbox}
\caption{XStoryCloze COMET translation metrics for different models.}
\label{tab:xstory_cloze_comet}
\end{table*}

\begin{table*}[ht]
\centering
\small
\begin{adjustbox}{max width=\textwidth}
\begin{tabular}{lrrrrrrrrrrrrr}
\toprule
Model &  Size &    et &    ht &    it &    id &    qu &    sw &    zh &    ta &    th &    tr &    vi &  avg \\
\midrule
 \multirow{4}{*}{NLLB} &   0.6B & 82.78 & 75.42 & 86.49 & 85.23 & 62.17 & 79.74 & 84.66 & 83.93 & 76.30 & 84.54 & 81.97 & 80.3 \\
  &   1.3B & 86.57 & 78.88 & 88.95 & 87.44 & 64.26 & 82.01 & 87.07 & 86.50 & 78.79 & 86.97 & 84.29 & 82.9 \\
  &   1.3B & 85.38 & 77.84 & 88.50 & 86.86 & 62.97 & 81.43 & 86.44 & 85.79 & 77.72 & 86.31 & 83.55 & 82.1 \\
  &   3.3B & 86.76 & 79.16 & 89.16 & 87.56 & 63.87 & 82.08 & 87.85 & 86.60 & 80.10 & 87.42 & 85.23 & 83.3 \\
 \midrule
 \multirow{4}{*}{XGLM} &   0.6B & 68.27 & 58.08 & 65.79 & 73.98 & 34.54 & 54.72 & 50.21 & 64.52 & 71.24 & 64.44 & 68.33 & 61.3 \\
  &   1.7B & 78.78 & 67.84 & 79.09 & 81.47 & 50.98 & 69.01 & 80.06 & 77.22 & 77.88 & 74.84 & 77.87 & 74.1 \\
  &   2.9B & 83.16 & 71.97 & 82.96 & 84.22 & 50.82 & 74.41 & 83.93 & 79.67 & 81.37 & 78.98 & 82.23 & 77.6 \\
  &   7.5B & 85.49 & 72.47 & 85.19 & 86.04 & 55.33 & 77.29 & 85.41 & 83.47 & 82.36 & 81.38 & 83.61 & 79.8 \\
 \midrule
\multirow{4}{*}{BLOOM} &   0.6B & 41.78 & 41.47 & 48.71 & 75.73 & 37.32 & 40.93 & 75.23 & 65.09 & 42.51 & 50.09 & 75.22 & 54.0 \\
 &   1.7B & 45.41 & 46.04 & 65.38 & 82.57 & 45.08 & 58.94 & 84.71 & 76.72 & 46.41 & 48.74 & 81.43 & 61.9 \\
 &   3B & 46.22 & 48.21 & 70.61 & 83.61 & 43.38 & 63.68 & 86.20 & 80.41 & 43.01 & 47.86 & 83.56 & 63.3 \\
 &   7.1B & 47.93 & 50.22 & 75.59 & 86.24 & 47.02 & 67.57 & 87.99 & 83.99 & 47.90 & 50.54 & 85.17 & 66.4 \\
\midrule
\multirow{3}{*}{LLaMA} &   7B & 51.26 & 48.89 & 85.89 & 70.59 & 49.65 & 50.03 & 80.04 & 49.16 & 53.79 & 59.32 & 54.76 & 59.4 \\
 &  13B & 52.17 & 49.01 & 87.22 & 75.13 & 48.00 & 50.14 & 83.16 & 49.02 & 58.65 & 67.93 & 59.71 & 61.8 \\
 &  30B & 55.41 & 52.29 & 88.42 & 79.85 & 48.48 & 54.73 & 85.10 & 52.96 & 59.66 & 71.51 & 66.20 & 65.0 \\
\bottomrule
\end{tabular}
\end{adjustbox}
\caption{XCOPA COMET translation metrics for different models.}
\label{tab:xcopa_comet}
\end{table*}

\begin{table*}[ht]
\centering
\small
\begin{adjustbox}{max width=\textwidth}
\begin{tabular}{lrrrrrrrrrrrrrrrr}
\toprule
Model &  Size &    ar &    bg &    de &    el &    es &    fr &    hi &    ru &    sw &    th &    tr &    ur &    vi &    zh &  avg \\
\midrule
 \multirow{4}{*}{NLLB} &   0.6B & 83.91 & 86.05 & 87.17 & 87.14 & 88.19 & 87.09 & 85.53 & 82.75 & 80.69 & 82.53 & 85.94 & 80.09 & 85.02 & 82.64 & 84.6 \\
  &   1.3B & 85.27 & 86.97 & 88.16 & 88.04 & 88.74 & 87.84 & 86.38 & 83.78 & 82.06 & 83.71 & 87.08 & 81.13 & 86.03 & 83.52 & 85.6 \\
  &   1.3B & 84.92 & 86.91 & 88.00 & 88.02 & 88.73 & 87.82 & 86.22 & 83.66 & 81.82 & 83.37 & 86.92 & 81.06 & 85.84 & 83.63 & 85.5 \\
  &   3.3B & 85.38 & 87.19 & 88.29 & 88.40 & 88.97 & 88.07 & 86.74 & 84.05 & 82.22 & 84.22 & 87.40 & 81.53 & 86.31 & 84.47 & 85.9 \\
 \midrule
 \multirow{4}{*}{XGLM} &   0.6B & 60.80 & 73.87 & 73.76 & 71.82 & 72.89 & 74.99 & 64.73 & 69.33 & 57.49 & 65.94 & 62.75 & 60.62 & 65.27 & 52.02 & 66.2 \\
  &   1.7B & 72.72 & 80.62 & 80.64 & 81.78 & 80.82 & 80.95 & 72.41 & 76.01 & 69.78 & 76.53 & 72.42 & 67.55 & 76.38 & 73.10 & 75.8 \\
  &   2.9B & 75.17 & 82.24 & 83.02 & 83.77 & 82.63 & 82.55 & 77.06 & 78.67 & 73.39 & 77.61 & 75.16 & 71.51 & 79.16 & 77.66 & 78.5 \\
  &   7.5B & 79.66 & 84.69 & 85.78 & 85.73 & 85.97 & 85.55 & 80.19 & 81.00 & 77.22 & 81.23 & 79.88 & 74.83 & 81.87 & 79.85 & 81.7 \\
 \midrule
\multirow{4}{*}{BLOOM} &   0.6B & 74.45 & 47.03 & 63.00 & 46.67 & 82.34 & 82.67 & 74.18 & 48.84 & 53.88 & 46.89 & 49.18 & 66.12 & 78.31 & 76.58 & 63.6 \\
 &   1.7B & 77.11 & 51.94 & 67.78 & 50.11 & 84.05 & 84.46 & 76.28 & 61.11 & 62.78 & 49.06 & 50.15 & 69.20 & 80.43 & 78.53 & 67.4 \\
 &   3B & 79.00 & 53.83 & 72.10 & 52.79 & 85.41 & 85.44 & 78.44 & 65.10 & 68.50 & 48.98 & 49.89 & 71.53 & 82.09 & 80.02 & 69.5 \\
 &   7.1B & 81.29 & 61.50 & 78.12 & 58.62 & 86.95 & 86.78 & 81.33 & 70.10 & 72.72 & 51.97 & 53.47 & 74.65 & 83.44 & 82.21 & 73.1 \\
\midrule
\multirow{3}{*}{LLaMA} &   7B & 66.76 & 83.89 & 86.57 & 72.61 & 86.94 & 86.65 & 66.69 & 81.54 & 51.36 & 58.09 & 64.03 & 54.27 & 62.59 & 78.32 & 71.5 \\
 &  13B & 72.16 & 85.07 & 87.45 & 77.56 & 87.82 & 87.32 & 72.59 & 82.65 & 53.52 & 63.76 & 72.12 & 59.76 & 68.36 & 80.35 & 75.0 \\
 &  30B & 77.03 & 86.36 & 88.14 & 82.33 & 88.32 & 87.78 & 78.50 & 83.40 & 60.13 & 66.14 & 76.34 & 67.02 & 74.72 & 81.74 & 78.4 \\
\bottomrule
\end{tabular}
\end{adjustbox}
\caption{XNLI COMET translation metrics for different models.}
\label{tab:xnli_comet}
\end{table*}

\begin{table*}[ht]
\centering
\small
\begin{adjustbox}{max width=\textwidth}
\begin{tabular}{lrrrrrrrr}
\toprule
Model &  Size &    de &    es &    fr &    ja &    ko &    zh &  avg \\
\midrule
 \multirow{4}{*}{NLLB} &   0.6B & 87.06 & 87.60 & 87.31 & 82.93 & 84.59 & 82.73 & 85.4 \\
  &   1.3B & 87.26 & 87.81 & 87.55 & 84.24 & 85.46 & 83.84 & 86.0 \\
  &   1.3B & 87.33 & 87.87 & 87.59 & 84.19 & 85.15 & 83.58 & 86.0 \\
  &   3.3B & 87.38 & 87.91 & 87.66 & 84.38 & 85.67 & 84.16 & 86.2 \\
 \midrule
 \multirow{4}{*}{XGLM} &   0.6B & 74.77 & 74.42 & 76.62 & 55.72 & 61.30 & 53.28 & 66.0 \\
  &   1.7B & 81.66 & 82.19 & 82.06 & 68.13 & 72.94 & 68.66 & 75.9 \\
  &   2.9B & 83.38 & 83.78 & 83.72 & 73.40 & 76.78 & 74.16 & 79.2 \\
  &   7.5B & 84.96 & 85.34 & 85.41 & 77.03 & 80.24 & 76.53 & 81.6 \\
 \midrule
\multirow{4}{*}{BLOOM} &   0.6B & 60.17 & 74.43 & 76.62 & 49.91 & 38.58 & 63.76 & 60.6 \\
 &   1.7B & 74.49 & 83.75 & 84.28 & 63.20 & 51.49 & 75.14 & 72.1 \\
 &   3B & 78.48 & 85.31 & 85.35 & 68.30 & 53.03 & 77.74 & 74.7 \\
 &   7.1B & 82.27 & 86.42 & 86.50 & 73.90 & 63.02 & 80.72 & 78.8 \\
\midrule
\multirow{3}{*}{LLaMA} &   7B & 85.97 & 86.47 & 86.16 & 76.41 & 75.19 & 74.98 & 80.9 \\
 &  13B & 86.28 & 86.77 & 86.65 & 79.96 & 78.81 & 77.40 & 82.6 \\
 &  30B & 86.64 & 87.26 & 86.99 & 81.35 & 81.29 & 79.34 & 83.8 \\
\bottomrule
\end{tabular}
\end{adjustbox}
\caption{PAWS-X COMET translation metrics for different models.}
\label{tab:pawsx_comet}
\end{table*}

\begin{table*}[ht]
\centering
\small
\begin{adjustbox}{max width=\textwidth}
\begin{tabular}{lrrrrrrrrrrrr}
\toprule
Model &  Size &    es &    fr &    de &    ru &    zh &    ja &    th &    sw &    bn &    te &  avg \\
\midrule
 \multirow{4}{*}{NLLB} &   0.6B & 83.35 & 81.43 & 83.48 & 78.24 & 79.93 & 77.46 & 75.73 & 77.38 & 82.09 & 83.17 & 80.2 \\
  &   1.3B & 85.87 & 84.95 & 86.28 & 82.53 & 81.98 & 83.34 & 78.59 & 82.22 & 86.59 & 85.94 & 83.8 \\
  &   1.3B & 85.47 & 84.44 & 85.72 & 81.47 & 82.34 & 84.20 & 78.43 & 82.18 & 86.18 & 84.72 & 83.5 \\
  &   3.3B & 86.11 & 85.03 & 86.31 & 82.37 & 83.50 & 84.37 & 80.86 & 83.11 & 86.98 & 86.46 & 84.5 \\
 \midrule
 \multirow{4}{*}{XGLM} &   0.6B & 61.85 & 63.52 & 66.69 & 58.59 & 52.41 & 50.28 & 52.25 & 45.19 & 49.66 & 46.16 & 54.7 \\
  &   1.7B & 77.49 & 74.92 & 77.79 & 71.00 & 64.53 & 64.92 & 68.06 & 63.58 & 58.97 & 62.38 & 68.4 \\
  &   2.9B & 81.03 & 79.37 & 81.37 & 77.40 & 69.27 & 74.94 & 70.80 & 71.23 & 65.38 & 64.14 & 73.5 \\
  &   7.5B & 83.08 & 81.77 & 83.00 & 79.92 & 77.53 & 79.17 & 77.06 & 76.18 & 77.61 & 77.03 & 79.2 \\
 \midrule
\multirow{4}{*}{BLOOM} &   0.6B & 64.35 & 64.33 & 42.94 & 34.70 & 61.24 & 40.60 & 32.91 & 37.54 & 56.54 & 47.12 & 48.2 \\
 &   1.7B & 71.25 & 74.20 & 64.94 & 51.54 & 72.33 & 59.10 & 41.21 & 52.78 & 68.19 & 61.26 & 61.7 \\
 &   3B & 83.14 & 83.27 & 72.70 & 61.37 & 77.96 & 66.53 & 42.30 & 61.34 & 74.30 & 67.71 & 69.1 \\
 &   7.1B & 85.39 & 84.36 & 78.50 & 66.82 & 82.18 & 74.39 & 43.42 & 70.81 & 82.77 & 76.45 & 74.5 \\
\midrule
\multirow{3}{*}{LLAMA} &   7B & 73.82 & 83.28 & 85.25 & 81.04 & 78.29 & 78.41 & 51.07 & 47.93 & 49.61 & 31.69 & 66.0 \\
 &  13B & 79.72 & 85.36 & 84.27 & 83.05 & 80.52 & 81.41 & 58.73 & 54.15 & 57.64 & 31.44 & 69.6 \\
 &  30B & 48.21 & 71.07 & 86.85 & 78.93 & 82.97 & 80.89 & 62.67 & 63.28 & 67.77 & 31.88 & 67.5 \\
\bottomrule
\end{tabular}
\end{adjustbox}
\caption{MGSM COMET translation metrics for different models.}
\label{tab:mgsm_comet}
\end{table*}

\begin{table*}[ht]
\centering
\small
\begin{adjustbox}{max width=\textwidth}
\begin{tabular}{lrrrrrrrrrrrr}
\toprule
Model & Size &     ru &     zh &     es &     ar &     hi &     id &     te &     sw &     eu &     my &    avg \\
\midrule
 \multirow{4}{*}{NLLB} &   0.6B & 40.98 & 30.04 & 47.98 & 49.46 & 45.07 & 38.44 & 29.45 & 41.51 & 35.24 & 22.00 & 38.0 \\
  &   1.3B & 44.12 & 30.57 & 50.52 & 53.09 & 48.62 & 40.98 & 32.19 & 43.86 & 33.77 & 28.18 & 40.6 \\
  &   1.3B & 43.22 & 32.07 & 50.42 & 52.91 & 48.08 & 41.13 & 31.39 & 44.17 & 35.63 & 29.94 & 40.9 \\
  &   3.3B & 44.59 & 34.80 & 51.33 & 54.80 & 49.16 & 42.27 & 33.09 & 45.00 & 33.55 & 29.69 & 41.8 \\
 \midrule
 \multirow{4}{*}{XGLM} &   0.6B & 15.67 &  1.54 & 14.36 &  6.16 &  7.52 & 16.92 &  1.28 &  3.82 &  2.81 &  0.67 &  7.1 \\
  &   1.7B & 25.62 & 16.08 & 28.64 & 21.40 & 16.22 & 26.07 & 10.46 & 21.17 & 11.38 &  7.94 & 18.5 \\
  &   2.9B & 29.08 & 21.68 & 36.22 & 26.32 & 24.91 & 28.86 & 11.37 & 27.19 & 20.04 & 12.40 & 23.8 \\
  &   7.5B & 34.40 & 25.20 & 40.85 & 34.45 & 30.32 & 33.59 & 17.05 & 33.48 & 23.33 & 16.84 & 29.0 \\
 \midrule
\multirow{4}{*}{BLOOM} &   0.6B &  0.37 &  9.67 & 20.55 & 14.70 &  9.94 & 19.55 &  1.93 &  0.43 &  1.96 &  0.11 &  7.9 \\
 &   1.7B &  9.03 & 22.26 & 35.84 & 26.14 & 18.45 & 27.74 &  9.01 & 12.67 & 11.56 &  0.06 & 17.3 \\
 &   3B & 11.42 & 25.12 & 39.51 & 28.93 & 22.60 & 29.62 & 11.11 & 18.32 & 15.80 &  0.07 & 20.2 \\
 &   7.1B & 16.37 & 30.53 & 43.21 & 35.44 & 31.19 & 34.16 & 15.07 & 23.71 & 22.27 &  0.10 & 25.2 \\
\midrule
\multirow{3}{*}{LLaMA} &   7B & 36.15 & 20.08 & 43.75 & 11.84 & 10.27 & 21.49 &  0.11 &  2.12 &  0.78 &  0.07 & 14.7 \\
 &  13B & 39.22 & 25.29 & 45.85 & 18.78 & 15.92 & 27.28 &  0.18 &  3.10 &  1.20 &  0.07 & 17.7 \\
 &  30B & 41.26 & 27.88 & 47.42 & 27.04 & 26.12 & 33.00 &  0.32 &  7.77 &  1.35 &  0.06 & 21.2 \\
\bottomrule
\end{tabular}
\end{adjustbox}
\caption{XStoryCloze BLEU translation metrics for different models.}
\label{tab:xstory_cloze_metrics}
\end{table*}

\begin{table*}[ht]
\centering
\small
\begin{adjustbox}{max width=\textwidth}
\begin{tabular}{lrrrrrrrrrrrrr}
\toprule
Model & Size &     et &     ht &     it &     id &     qu &     sw &     zh &     ta &     th &     tr &     vi &    avg \\
\midrule
 \multirow{4}{*}{NLLB} &   0.6B & 39.07 & 33.85 & 45.88 & 33.15 &  9.26 & 32.29 & 35.16 & 32.33 & 21.23 & 37.66 & 32.81 & 32.1 \\
  &   1.3B & 45.42 & 40.40 & 51.01 & 37.41 & 12.02 & 35.57 & 38.20 & 37.47 & 24.75 & 42.61 & 37.47 & 36.6 \\
  &   1.3B & 43.75 & 38.26 & 50.93 & 37.22 & 10.48 & 35.39 & 38.52 & 37.36 & 23.36 & 40.93 & 35.67 & 35.6 \\
  &   3.3B & 45.57 & 40.42 & 52.45 & 38.12 & 11.38 & 36.91 & 42.42 & 38.34 & 26.36 & 43.06 & 38.90 & 37.6 \\
 \midrule
 \multirow{4}{*}{XGLM} &   0.6B & 12.08 &  9.37 & 10.06 & 12.99 &  0.35 &  2.96 &  0.92 &  2.29 &  7.67 &  4.62 &  8.73 &  6.5 \\
  &   1.7B & 25.29 & 20.36 & 28.12 & 23.88 &  1.16 & 15.62 & 22.94 & 12.69 & 12.80 & 15.54 & 20.31 & 18.1 \\
  &   2.9B & 34.93 & 25.21 & 32.88 & 27.51 &  1.91 & 21.70 & 29.21 & 17.77 & 22.52 & 22.32 & 29.36 & 24.1 \\
  &   7.5B & 39.55 & 28.41 & 40.18 & 31.90 &  4.11 & 27.25 & 32.50 & 25.27 & 24.79 & 26.41 & 32.14 & 28.4 \\
 \midrule
\multirow{4}{*}{BLOOM} &   0.6B &  0.09 &  0.22 &  2.40 & 16.07 &  0.17 &  0.11 & 13.70 &  4.35 &  0.08 &  0.10 & 15.63 &  4.8 \\
 &   1.7B &  0.24 &  0.59 & 13.94 & 25.17 &  0.37 &  6.59 & 28.91 & 12.37 &  0.08 &  0.20 & 27.26 & 10.5 \\
 &   3B &  0.29 &  1.39 & 19.83 & 27.15 &  0.31 & 10.67 & 34.77 & 18.77 &  0.13 &  0.20 & 29.82 & 13.0 \\
 &   7.1B &  0.76 &  2.88 & 26.80 & 32.87 &  0.48 & 15.72 & 39.41 & 26.92 &  0.18 &  0.70 & 34.91 & 16.5 \\
\midrule
\multirow{3}{*}{LLaMA} &   7B &  2.02 &  1.55 & 41.18 & 15.44 &  0.59 &  1.00 & 25.01 &  0.16 &  1.86 &  5.15 &  3.98 &  8.9 \\
 &  13B &  3.19 &  3.10 & 44.11 & 22.01 &  0.54 &  1.49 & 32.41 &  0.14 &  6.06 & 14.36 &  8.48 & 12.4 \\
 &  30B &  5.67 &  5.67 & 48.64 & 26.64 &  1.10 &  5.20 & 35.41 &  0.68 &  6.62 & 18.91 & 14.96 & 15.4 \\
\bottomrule
\end{tabular}
\end{adjustbox}
\caption{XCOPA BLEU translation metrics for different models.}
\label{tab:xcopa_metrics}
\end{table*}

\begin{table*}[ht]
\centering
\small
\begin{adjustbox}{max width=\textwidth}
\begin{tabular}{lrrrrrrrrrrrrrrrr}
\toprule
Model & Size &     ar &     bg &     de &     el &     es &     fr &     hi &     ru &     sw &     th &     tr &     ur &     vi &     zh &    avg \\
\midrule
\multirow{4}{*}{NLLB} &   0.6B & 37.99 & 41.39 & 44.65 & 46.13 & 50.92 & 45.09 & 38.09 & 31.41 & 34.09 & 28.16 & 36.28 & 30.61 & 39.10 & 27.71 & 38.0 \\
  &   1.3B & 41.09 & 43.80 & 46.97 & 48.54 & 53.02 & 47.17 & 40.78 & 33.49 & 36.30 & 30.00 & 39.24 & 32.84 & 41.81 & 29.48 & 40.3 \\
  &   1.3B & 40.56 & 43.62 & 46.69 & 48.37 & 53.05 & 46.81 & 40.40 & 33.36 & 36.45 & 29.90 & 39.00 & 32.28 & 41.41 & 29.52 & 40.1 \\
  &   3.3B & 42.19 & 45.08 & 47.66 & 50.05 & 53.80 & 47.73 & 41.73 & 33.98 & 37.89 & 31.35 & 40.61 & 33.86 & 43.20 & 31.31 & 41.5 \\
 \midrule
 \multirow{4}{*}{XGLM} &   0.6B &  5.54 & 17.83 & 19.91 & 14.67 & 17.56 & 20.52 &  5.91 & 12.07 &  4.97 &  7.25 &  4.38 &  4.50 &  8.85 &  1.67 & 10.4 \\
  &   1.7B & 16.34 & 27.20 & 30.30 & 30.86 & 31.54 & 29.73 & 12.77 & 18.83 & 16.63 & 15.23 & 11.78 &  9.81 & 21.11 & 12.36 & 20.3 \\
  &   2.9B & 19.63 & 30.91 & 34.54 & 35.14 & 34.76 & 32.98 & 17.96 & 22.45 & 20.83 & 17.68 & 15.09 & 13.58 & 24.71 & 16.84 & 24.1 \\
  &   7.5B & 26.52 & 35.23 & 38.80 & 39.16 & 41.56 & 38.93 & 22.09 & 25.91 & 26.29 & 22.56 & 19.71 & 17.61 & 29.08 & 19.80 & 28.8 \\
 \midrule
\multirow{4}{*}{BLOOM} &   0.6B & 17.71 &  1.35 & 12.21 &  1.08 & 33.99 & 33.08 & 12.62 &  2.10 &  4.35 &  0.92 &  0.90 &  7.53 & 22.30 & 14.71 & 11.8 \\
 &   1.7B & 21.61 &  3.34 & 16.19 &  2.71 & 37.73 & 36.64 & 15.36 &  8.77 & 10.58 &  1.07 &  1.21 & 10.26 & 26.12 & 16.82 & 14.9 \\
 &   3B & 24.10 &  4.43 & 19.05 &  4.42 & 40.60 & 38.84 & 17.61 & 11.22 & 15.99 &  1.48 &  1.35 & 12.46 & 28.96 & 19.12 & 17.1 \\
 &   7.1B & 29.03 &  9.79 & 28.06 &  8.66 & 45.07 & 42.44 & 22.74 & 15.50 & 21.16 &  2.53 &  3.08 & 16.73 & 31.94 & 23.17 & 21.4 \\
\midrule
\multirow{3}{*}{LLaMA} &   7B & 12.20 & 34.86 & 40.86 & 21.27 & 45.28 & 41.66 &  8.71 & 27.39 &  4.21 &  4.52 &  7.48 &  2.47 &  9.31 & 18.84 & 19.9 \\
 &  13B & 18.52 & 37.83 & 43.71 & 28.47 & 47.70 & 44.06 & 14.83 & 29.60 &  5.95 &  8.62 & 14.10 &  5.78 & 15.83 & 21.96 & 24.1 \\
 &  30B & 23.77 & 40.77 & 45.77 & 35.73 & 49.45 & 45.64 & 21.00 & 31.00 &  9.46 &  9.96 & 18.75 & 10.62 & 21.48 & 24.90 & 27.7 \\
\bottomrule
\end{tabular}
\end{adjustbox}
\caption{XNLI BLEU translation metrics for different models.}
\label{tab:xnli_metrics}
\end{table*}

\begin{table*}[ht]
\centering
\small
\begin{tabular}{lrrrrrrrr}
\toprule
Model & Size &     de &     es &     fr &     ja &     ko &     zh &    avg \\
\midrule
 \multirow{4}{*}{NLLB} &   0.6B & 59.41 & 64.80 & 61.18 & 33.09 & 38.52 & 36.94 & 49.0 \\
  &   1.3B & 60.52 & 65.56 & 62.66 & 37.53 & 41.48 & 40.08 & 51.3 \\
  &   1.3B & 60.66 & 65.72 & 62.52 & 36.80 & 40.77 & 38.89 & 50.9 \\
  &   3.3B & 61.19 & 66.02 & 62.91 & 38.12 & 41.97 & 41.21 & 51.9 \\
 \midrule
 \multirow{4}{*}{XGLM} &   0.6B & 30.41 & 31.70 & 34.00 &  2.89 &  5.64 &  3.42 & 18.0 \\
  &   1.7B & 44.35 & 47.33 & 43.03 &  9.13 & 14.64 & 11.34 & 28.3 \\
  &   2.9B & 48.69 & 51.59 & 48.39 & 14.21 & 19.19 & 16.79 & 33.1 \\
  &   7.5B & 51.22 & 54.58 & 53.12 & 18.27 & 24.89 & 20.09 & 37.0 \\
 \midrule
\multirow{4}{*}{BLOOM} &   0.6B & 15.95 & 33.98 & 34.67 &  2.79 &  1.06 &  8.69 & 16.2 \\
 &   1.7B & 32.25 & 50.68 & 49.56 &  7.38 &  5.61 & 17.85 & 27.2 \\
 &   3B & 39.59 & 54.56 & 53.02 & 11.09 &  6.83 & 21.66 & 31.1 \\
 &   7.1B & 45.61 & 58.41 & 56.59 & 15.89 & 12.61 & 27.48 & 36.1 \\
\midrule
\multirow{3}{*}{LLaMA} &   7B & 56.24 & 59.61 & 56.48 & 20.55 & 21.77 & 19.70 & 39.1 \\
 &  13B & 57.36 & 61.05 & 58.86 & 26.16 & 26.98 & 24.52 & 42.5 \\
 &  30B & 59.61 & 63.07 & 60.47 & 30.07 & 31.75 & 27.48 & 45.4 \\
\bottomrule
\end{tabular}
\caption{PAWS-X BLEU translation metrics for different models.}
\label{tab:pawsx_metrics}
\end{table*}

\begin{table*}[ht]
\centering
\small
\begin{adjustbox}{max width=\textwidth}
\begin{tabular}{lrrrrrrrrrrrr}
\toprule
Model & Size &     es &     fr &     de &     ru &     zh &     ja &     th &     sw &     bn &     te &    avg \\
\midrule
 \multirow{4}{*}{NLLB} &   0.6B & 48.34 & 34.85 & 44.57 & 31.39 & 28.14 & 17.99 & 17.37 & 34.62 & 28.58 & 34.68 & 32.1 \\
  &   1.3B & 57.94 & 44.44 & 54.21 & 45.11 & 33.23 & 29.69 & 19.62 & 46.91 & 40.80 & 41.54 & 41.3 \\
  &   1.3B & 56.78 & 44.00 & 52.64 & 42.11 & 33.91 & 33.51 & 19.83 & 47.51 & 39.82 & 38.45 & 40.9 \\
  &   3.3B & 57.91 & 44.26 & 53.41 & 44.85 & 38.44 & 35.59 & 24.30 & 51.37 & 42.89 & 44.02 & 43.7 \\
 \midrule
 \multirow{4}{*}{XGLM} &   0.6B & 12.94 & 11.30 & 15.94 &  7.53 &  1.77 &  0.82 &  1.22 &  1.27 &  0.77 &  0.60 &  5.4 \\
  &   1.7B & 36.77 & 24.31 & 33.33 & 23.89 &  8.26 &  6.14 &  9.32 & 16.76 &  5.43 &  6.50 & 17.1 \\
  &   2.9B & 44.50 & 32.70 & 40.77 & 33.20 & 13.25 & 14.41 & 10.71 & 24.70 & 11.80 &  9.28 & 23.5 \\
  &   7.5B & 45.04 & 33.37 & 41.55 & 34.70 & 20.75 & 20.09 & 18.44 & 31.32 & 19.11 & 18.63 & 28.3 \\
 \midrule
\multirow{4}{*}{BLOOM} &   0.6B & 19.40 & 13.29 &  4.75 &  0.38 &  7.83 &  1.14 &  0.06 &  0.67 &  4.33 &  1.97 &  5.4 \\
 &   1.7B & 28.14 & 25.34 & 17.91 &  9.39 & 15.72 &  5.40 &  0.14 &  7.56 &  9.10 &  7.23 & 12.6 \\
 &   3B & 47.91 & 37.39 & 27.37 & 16.90 & 22.32 &  9.92 &  0.08 & 15.02 & 15.92 & 10.25 & 20.3 \\
 &   7.1B & 54.44 & 41.80 & 35.30 & 23.42 & 29.46 & 15.98 &  0.36 & 29.03 & 27.69 & 19.46 & 27.7 \\
\midrule
\multirow{3}{*}{LLaMA} &   7B & 44.51 & 41.92 & 51.04 & 43.48 & 25.82 & 20.86 &  2.86 &  5.77 &  3.02 &  0.00 & 23.9 \\
 &  13B & 53.27 & 44.99 & 52.85 & 47.92 & 29.82 & 26.69 &  6.26 &  9.66 &  7.61 &  0.00 & 27.9 \\
 &  30B & 14.17 & 33.08 & 56.09 & 45.29 & 35.58 & 30.84 &  8.40 & 17.40 & 14.19 &  0.00 & 25.5 \\
\bottomrule
\end{tabular}
\end{adjustbox}
\caption{MGSM BLEU translation metrics for different models.}
\label{tab:mgsm_metrics}
\end{table*}

\end{document}